\documentclass[10pt]{article}

\usepackage[paper=a4paper,top=18mm,bottom=20mm,left=20mm,right=20mm]{geometry}
\usepackage{amsmath,amssymb}
\usepackage{etoolbox}
\usepackage{latexml}
\usepackage{fontspec}
\usepackage{graphicx}
\usepackage{booktabs}
\usepackage[table]{xcolor}
\usepackage{caption}
\usepackage{float}
\usepackage{titlesec}
\usepackage{fancyhdr}
\usepackage{enumitem}
\usepackage{multicol}
\usepackage{balance}
\usepackage{ragged2e}
\usepackage[skip=0.55em plus 0.1em, indent=0pt]{parskip}
\usepackage[colorlinks=true, linkcolor=ink, citecolor=g500, urlcolor=ink]{hyperref}

\AtBeginDocument{\appto\normalsize{%
  \setlength{\abovedisplayskip}{10pt plus 2pt minus 2pt}%
  \setlength{\belowdisplayskip}{10pt plus 2pt minus 2pt}%
  \setlength{\abovedisplayshortskip}{8pt plus 2pt}%
  \setlength{\belowdisplayshortskip}{8pt plus 2pt minus 2pt}}}

\definecolor{ink}{HTML}{0A0A0A}
\definecolor{g50}{HTML}{F7F7F7}
\definecolor{g100}{HTML}{EEEEEE}
\definecolor{g200}{HTML}{DDDDDD}
\definecolor{g300}{HTML}{BEBEBE}
\definecolor{g400}{HTML}{909090}
\definecolor{g500}{HTML}{6B6B6B}
\definecolor{g600}{HTML}{4A4A4A}
\definecolor{bestcell}{HTML}{EEEEEE}
\definecolor{runnerupcell}{HTML}{F7F7F7}
\newcommand{\best}[1]{\cellcolor{bestcell}\textbf{#1}}

\arrayrulecolor{g200}

\newfontfamily\inter[BoldFont=Inter-SemiBold.ttf]{Inter-Regular.ttf}
\newfontfamily\intermed{Inter-Medium.ttf}
\newfontfamily\brock[BoldFont=Brockmann-Bold.otf,ItalicFont=Brockmann-SemiBoldItalic.otf,LetterSpace=-2.5]{Brockmann-SemiBold.otf}
\providecommand{\inter}{}\providecommand{\intermed}{}\providecommand{\brock}{}
\providecommand{\addfontfeature}[1]{}
\newcommand{\eyebrow}[1]{{\intermed\fontsize{7.2}{9}\selectfont\addfontfeature{LetterSpace=14}\color{g500}\MakeUppercase{#1}}}
\newcommand{\eyebrowmono}[1]{{\ttfamily\fontsize{6.8}{9}\selectfont\addfontfeature{LetterSpace=14}\color{g500}\MakeUppercase{#1}}}

\newcommand{\twodigits}[1]{\ifnum#1<10 0\fi\number#1}

\renewcommand{\thesubsection}{\arabic{section}.\arabic{subsection}}
\iflatexml\else
\titleformat{\section}{\brock\fontsize{15.5}{18}\selectfont\color{ink}}{\twodigits{\value{section}}\;·\;}{0pt}{}
\titlespacing*{\section}{0pt}{1.9em plus 0.3em}{0.55em}
\titleformat{\subsection}{\inter\bfseries\fontsize{10.5}{13}\selectfont\color{ink}}{\thesubsection}{0.5em}{}
\titlespacing*{\subsection}{0pt}{1.3em plus 0.2em}{0.3em}
\titleformat{\paragraph}[runin]{\inter\bfseries\color{ink}}{}{0pt}{}
\titlespacing*{\paragraph}{0pt}{0.7em}{0.45em}
\fi

\iflatexml\else
\DeclareCaptionLabelFormat{sl}{{\eyebrowmono{#1 #2}}}
\DeclareCaptionFormat{slfig}{\noindent{\color{g200}\rule{\linewidth}{0.4pt}}\par\vspace{4pt}\noindent#1\hspace{0.7em}#3\par}
\DeclareCaptionFormat{sltab}{\noindent#1\hspace{0.7em}#3\par\vspace{3pt}}
\fi

\newcommand{\thd}[1]{{\ttfamily\fontsize{6.8}{9}\selectfont\addfontfeature{LetterSpace=14}\color{g500}\MakeUppercase{#1}}}
\renewcommand{\toprule}{}
\renewcommand{\midrule}{\specialrule{1.4pt}{2pt}{4pt}\arrayrulecolor{g200}}
\renewcommand{\bottomrule}{\specialrule{0.4pt}{2pt}{0pt}}

\AtBeginEnvironment{tabular}{\small\arrayrulecolor{ink}}

\fancypagestyle{first}{\fancyhf{}\fancyfoot[L]{\eyebrowmono{Preprint · 2026 · Speridlabs Research}}\fancyfoot[R]{\eyebrowmono{IRIS TECHNICAL REPORT · \thepage}}}
\setlist[itemize]{leftmargin=1.2em,itemsep=0.3em,label={\color{g400}\rule[0.5ex]{0.5em}{0.5pt}}}

\newsavebox{\abstractbox}
\newlength{\abstractH}
\iflatexml\else
\renewenvironment{abstract}{%
  \begin{lrbox}{\abstractbox}\begin{minipage}{\dimexpr\linewidth-14.2mm\relax}%
  \vspace{4mm}\setlength{\parskip}{0.5em}\small\linespread{1.22}\selectfont%
  \noindent\eyebrow{Abstract}\par\vspace{0.6em}}%
  {\par\vspace{4mm}\end{minipage}\end{lrbox}%
  \setlength{\abstractH}{\dimexpr\ht\abstractbox+\dp\abstractbox\relax}%
  \noindent{\setlength{\fboxsep}{0pt}\setlength{\fboxrule}{0.4pt}%
  \fcolorbox{g200}{g50}{\raisebox{-\dp\abstractbox}{\color{ink}\rule{2.2pt}{\abstractH}}\hspace{6mm}\usebox{\abstractbox}\hspace{6mm}}}\par\vspace{6mm}}
\fi

\newcommand{\preprinttitle}{%
  \setlength{\parskip}{0pt}%
  \noindent\begin{minipage}[b]{0.5\linewidth}\raisebox{-0.35\height}{\includegraphics[height=6.2mm]{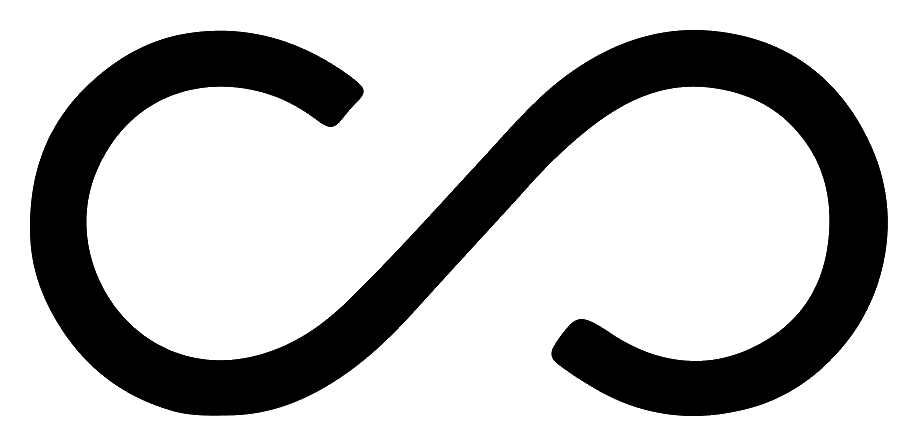}}\hspace{2.2mm}{\inter\fontsize{12}{14}\selectfont\color{ink}speridlabs}\end{minipage}%
  \begin{minipage}[b]{0.5\linewidth}\raggedleft\eyebrowmono{Preprint · Technical report}\\[2pt]\eyebrowmono{2026 · Speridlabs Research}\end{minipage}\par
  \vspace{3mm}\noindent{\color{ink}\rule{\linewidth}{1.4pt}}\par\vspace{7mm}%
  \noindent\eyebrow{Pixel-space generative models}\par\vspace{2.5mm}%
  \begin{flushleft}\brock\fontsize{25}{27}\selectfont\color{ink}Iris-3B: Going Beyond the Latent with Pixel-Space Diffusion Training, Conversion and Fine-Tuning\end{flushleft}\vspace{2.5mm}%
  \begin{flushleft}\inter\fontsize{10.5}{14}\selectfont\color{ink}%
  \textbf{Hanqiu Li Cai}\textsuperscript{\textdagger}, \textbf{Chema Garabito}\\[0.5em]
  {\footnotesize\color{g500}SperidLabs\;·\;\textsuperscript{\textdagger}Project lead\;·\;Code and models: \href{https://github.com/speridlabs/iris-3b}{\color{ink}github.com/speridlabs/iris-3b}}\end{flushleft}\vspace{2mm}%
  \setlength{\parskip}{0.55em plus 0.1em}}

\makeatletter
\iflatexml\else
\renewcommand\@biblabel[1]{{\ttfamily\fontsize{7}{9}\selectfont\color{g500}[\ifnum#1<10 0\fi#1]}}
\renewenvironment{thebibliography}[1]{%
  \section*{References}\addcontentsline{toc}{section}{References}%
  \fontsize{8}{10.5}\selectfont\color{g600}%
  \list{\@biblabel{\@arabic\c@enumiv}}{\settowidth\labelwidth{\@biblabel{#1}}%
  \leftmargin\labelwidth\advance\leftmargin\labelsep\itemsep 0.1em\parsep 0pt\usecounter{enumiv}}%
  \sloppy\clubpenalty4000\widowpenalty4000}{\endlist}
\fi
\makeatother

\title{Iris-3B: Going Beyond the Latent with Pixel-Space Diffusion Training, Conversion and Fine-Tuning}
\author{Hanqiu Li Cai\textsuperscript{\textdagger} \and Chema Garabito\\
SperidLabs \quad \textsuperscript{\textdagger}Project lead \quad Code and models:
\href{https://github.com/speridlabs/iris-3b}{github.com/speridlabs/iris-3b}}
\date{}
\iflatexml
\renewcommand{\thd}[1]{#1}\renewcommand{\eyebrowmono}[1]{#1}
\renewcommand{\eyebrow}[1]{#1}
\fi

\begin{document}
\iflatexml\maketitle\else\preprinttitle\thispagestyle{first}\fi

\begin{abstract}
Pixel-space diffusion models avoid the lossy VAE of latent models, which suggests an advantage on
downstream tasks where fine-grained detail matters. We test this claim along both routes to a
pixel-space backbone. We pretrain \textbf{Iris-3B}, a 3B-parameter pixel-space text-to-image
transformer, from scratch through a 256\,$\rightarrow$\,512\,$\rightarrow$\,1024 curriculum, after
first ablating the prediction target and representation alignment (REPA) at $256^2$ to decide what to
scale. We also convert a pretrained latent model, FLUX.2 Klein base 4B~\cite{flux2}, to pixel space. We fine-tune
both families for monocular depth estimation and for image restoration/super-resolution. We find no
significant improvement from using a pixel-space generative prior. Fine-tuned for depth with one
matched direct-regression recipe, Iris-3B is level with the latent FLUX.2 Klein and the converted
pixel FLUX.2 Klein falls behind it, and on $4\times$ DIV2K restoration neither pixel model beats a
latent FLUX.2 Klein fine-tune, the converted one trailing it slightly. We document the recipes, the failure modes and the
remaining confounds behind this negative result. Nevertheless, Iris-3B shows that pixel-space
pretraining with the pixel-transformer (PiT) head of PixelDiT~\cite{pixeldit} scales to 3B
parameters and to text-to-image quality competitive with latent models, matching
Qwen-Image~\cite{qwenimage} on OneIG under the official evaluators at $1024^2$. We
release its weights and training code in the hope that they help pave the way for further work on
pixel-space generation.
\end{abstract}

{\centering\includegraphics[width=\linewidth,height=\dimexpr\pagegoal-\pagetotal-9mm\relax,keepaspectratio]{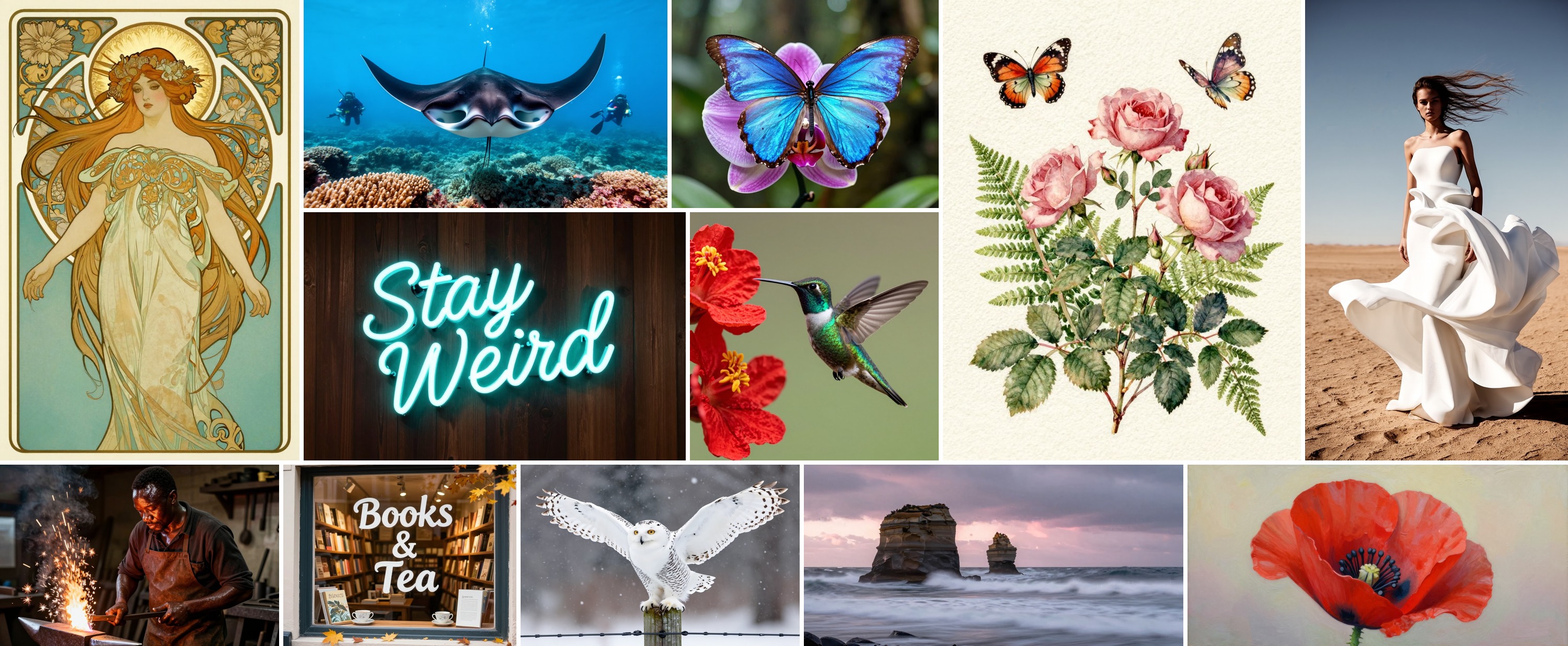}\par}\vspace{1mm}
\noindent{\footnotesize\color{g500}\eyebrowmono{Fig. 1}\hspace{0.7em}Samples from Iris-3B at a
one-megapixel budget across aspect ratios, generated directly in pixel space.}\par\vspace{5mm}
\setcounter{figure}{1}

\begin{multicols}{2}

\section{Introduction}
\label{sec:intro}
Latent diffusion~\cite{ldm} has become the dominant paradigm for high-resolution image synthesis.
By running the generative process in the compressed space of a pretrained variational autoencoder
(VAE), it makes training and sampling at high resolution affordable. This efficiency comes at a
price. The autoencoder is trained separately from the generator and reconstructs with loss, so its
errors compound with those of the generator and the two cannot be optimized
jointly~\cite{pixeldit}; any detail the codec discards lies beyond the generator's reach.
Generating directly in pixel space removes this bottleneck, and recent work has made it
competitive: transformers operating on raw pixels now rival latent models on class-conditional
ImageNet~\cite{jit} and approach them in text-to-image quality~\cite{pixeldit,asymflow},
pretrained latent models can be converted to pixel space
at a fraction of the cost of training from scratch~\cite{l2p,empiricalstudy}, and raw pixel
patches can serve as the only visual interface of a unified model that both understands and
generates images and video~\cite{pixelumm}.

These works, however, judge pixel-space models by what they generate or understand, not by how
well they serve as backbones for other tasks. Pretrained generators are increasingly valued for
exactly that reason, as backbones for other
vision tasks: fine-tuned generators define the state of the art in monocular depth
estimation~\cite{marigold,marigoldv2} and supply strong priors for image
restoration~\cite{hypir}. For such tasks the codec is a structural limitation rather than an
aesthetic one. When the backbone is a latent model, every dense output, whether a depth map or a
restored photograph, must be decoded by an autoencoder that was optimized to reconstruct natural
images. It is therefore natural to expect that \emph{a pixel-space generator should transfer
better than a latent one to tasks in which fine spatial detail matters.}

In this report we put this hypothesis to the test, obtaining pixel-space backbones by both routes
the literature offers. First, we pretrain Iris-3B, a 3B-parameter pixel-space text-to-image
transformer, from scratch through a 256\,$\rightarrow$\,512\,$\rightarrow$\,1024 curriculum. To
choose its recipe, we ablate pretraining strategies on a 1.3B PixelDiT at $256^2$: prediction
targets ($x$ vs.\ $v$) and representation-alignment objectives (REPA, iREPA, DINOv3 targets and
Self-Flow), each compared with its own control at matched steps (Sec.~\ref{sec:ablations}).
Under the official evaluators at $1024^2$, Iris-3B reaches GenEval 0.798 and DPG 86.52, showing
that pixel-space pretraining is practical at this scale. Second, we convert a pretrained latent
model, FLUX.2 Klein base 4B~\cite{flux2}, to pixel space. We then fine-tune both kinds of model for two
detail-critical tasks, monocular depth estimation and image restoration with super-resolution, and
compare them against latent counterparts.

Contrary to our expectation, we find no significant improvement from a pixel-space generative
prior. We fine-tune Iris-3B, the converted FLUX.2 Klein and the latent FLUX.2 Klein for depth with
one matched recipe and score them on NYUv2, KITTI, ETH3D, ScanNet and DIODE: Iris-3B is level with
the latent model, and the converted pixel model falls behind it. On restoration, the converted
model trails its latent counterpart slightly. None of these
comparisons is free of confounds, and we document
those that remain rather than claim to have resolved them. We nonetheless believe that the
negative result, together with the recipes that produced it, is useful to anyone weighing
pixel-space backbones for dense prediction.

In summary, we make the following contributions:
\begin{itemize}
\item We explore fine-tuning pixel-space generative models, both trained from scratch and
converted from latent models, for monocular depth estimation and image restoration, and evaluate
them against latent counterparts. We find no significant improvement from pixel space and document the confounds
that remain (Sec.~\ref{sec:downstream}).
\item We evaluate pretraining strategies for pixel-space models in a controlled 256px ablation
campaign, covering prediction targets ($x$ vs.\ $v$) and representation-alignment strategies (REPA,
iREPA, DINOv3 targets and Self-Flow) (Sec.~\ref{sec:ablations}).
\item We convert FLUX.2 Klein base 4B to pixel space and identify the input interface as the
obstacle, which a gain-matched ridge initialization and a scaled trunk learning rate (LR) remove
(Sec.~\ref{sec:l2p}).
\item We release \textbf{Iris-3B}, an open 3B-parameter pixel-space text-to-image model
trained from scratch, showing that pixel-space generation works at scale, together with its
weights and training code.
\end{itemize}

\section{Related Work}
\label{sec:related}

\begin{figure*}[b]
\includegraphics[width=\linewidth]{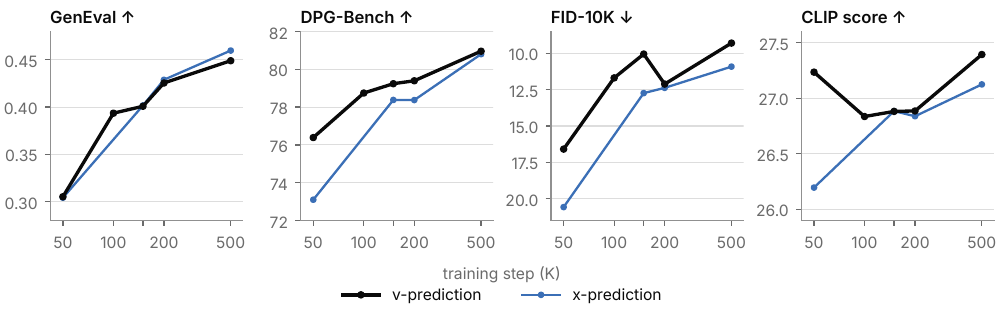}
\caption{$x$- versus $v$-prediction at $256^2$, absolute scores at matched steps
(EMA, 25-step FlowDPM++, CFG 4.5). Following JiT~\cite{jit}, the $x$-prediction arm also uses its
logit-normal timestep distribution $(0.8, 0.8)$ instead of $(0.0, 1.0)$, so the contrast does not
isolate the prediction target.}
\label{fig:prediction}
\end{figure*}

\paragraph{Pixel-space generation.}
Early pixel-space diffusion models relied on cascades or on pixel-level U-Nets. Recent work shows
that a single transformer can generate in pixel space. JiT~\cite{jit} uses large patches and
predicts clean images, which avoids the failure of noise prediction in high dimensions.
PixelDiT~\cite{pixeldit} splits the model into a patch-level transformer and a pixel-level
transformer that refines texture, and scales to high-resolution text-to-image generation.
DiP~\cite{dip} and PixNerd~\cite{pixnerd} replace the per-patch linear decoder with a lightweight
pixel decoder and a neural field, respectively. PixelGen~\cite{pixelgen} adds perceptual losses to
clean-image prediction and trains a competitive text-to-image model on a small compute budget.
AsymFlow~\cite{asymflow} restricts noise prediction to a low-rank subspace. Concurrent with this
report, PixelUMM~\cite{pixelumm} extends the encoder-free approach to a unified model that
understands and generates images and video in pixel space. Raw pixel patches are its only visual
interface, with no ViT encoder and no VAE. Its design studies anticipate several of our own
findings, which we note where they arise.

\paragraph{Latent-to-pixel conversion.}
A complementary line of work obtains pixel-space models by converting pretrained latent ones.
L2P~\cite{l2p} replaces the VAE with large-patch tokenization, freezes the intermediate layers and
trains only the shallow ones on images generated by the source model, approaching the quality of
the source at a small fraction of its training cost. Jiang et al.~\cite{empiricalstudy} study this
transition systematically, ablating initialization, data composition, prediction target, decoder
and noise schedule, and obtain converted models that match or exceed their latent counterparts
while running faster end to end. AsymFlow~\cite{asymflow} instead aligns its low-rank pixel
subspace with the latent space, which yields a seamless initialization from FLUX.2 Klein 9B and a
converted model that surpasses its latent base.

\paragraph{Generative priors for dense prediction.}
Vision Banana~\cite{visionbanana} argues that generation pretraining yields generalist visual
representations. Marigold~\cite{marigold} fine-tunes a latent text-to-image model into an affine-invariant depth
estimator, and E2E-FT~\cite{e2eft} shows that a single-step, end-to-end fine-tune works as well as
the multi-step original; our depth recipe follows this single-step design. Marigold
V2~\cite{marigoldv2} moves to modern DiTs, adds representation alignment against ground-truth
features and a Sinkhorn-based second stage; it names the recovery of sharp, detailed depth among the open problems of
the field. PointDiT~\cite{pointdit} trains a pixel-space DiT for point maps from scratch,
conditioned on DINOv3 features~\cite{dinov3}, and shows that a point-map VAE injects reconstruction
noise whereas denoising raw point-map patches recovers sharper structure.

\paragraph{Generative restoration.}
Diffusion-based restoration conditions a pretrained text-to-image generator on the low-quality
image. StableSR~\cite{stablesr} adds a time-aware encoder to Stable Diffusion, and
DiffBIR~\cite{diffbir} first removes degradations with a restoration module and then regenerates
detail with a diffusion prior. SeeSR~\cite{seesr} adds semantic prompts, and SUPIR~\cite{supir}
scales the approach to SDXL with large-scale training data and text guidance. To reduce the cost of
iterative sampling, OSEDiff~\cite{osediff} distills restoration
into a single step with variational score distillation, and HYPIR~\cite{hypir} initializes a
one-step restoration network from a diffusion model's score prior and trains it adversarially.
Training degradations typically follow Real-ESRGAN~\cite{realesrgan}.

\paragraph{Representation alignment.}
REPA~\cite{repa} aligns intermediate DiT features with a frozen DINOv2~\cite{dinov2} encoder and
speeds up training substantially. iREPA~\cite{irepa} argues that the spatial structure of the
target features matters more than their global content. Self-Flow~\cite{selfflow} replaces the
external encoder with a self-supervised objective built into flow matching. We ablate all three
(Sec.~\ref{sec:ablations}).

\section{Iris-3B: Pretraining a Pixel-Space Model}
\label{sec:pretrain}

\subsection{Ablations at 256px}
\label{sec:ablations}
\begin{figure*}[!b]
\includegraphics[width=\linewidth]{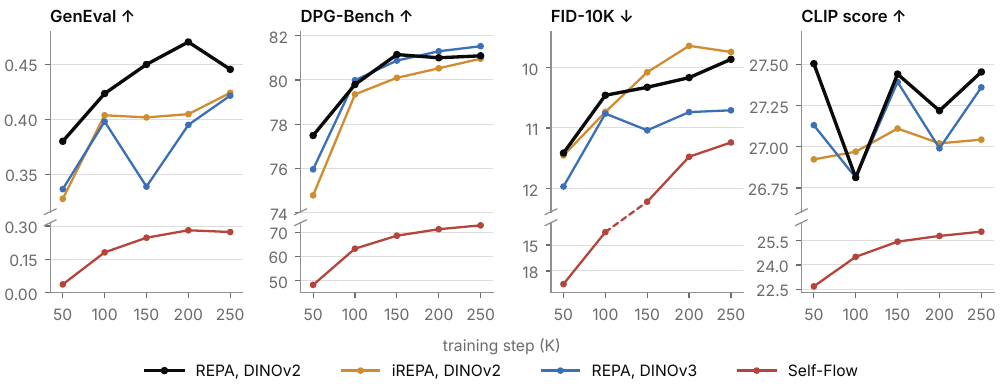}
\caption{Representation-alignment ablations at $256^2$, absolute scores at
matched steps (same protocol as Fig.~\ref{fig:prediction}). All arms share the same model and data
and differ only in the alignment objective. Each y-axis is broken to fit Self-Flow on the
same scale.}
\label{fig:ablation}
\end{figure*}
We first select the pretraining recipe with ablations on a 1.3B dual-stream PixelDiT
reproduction~\cite{pixeldit} at $256^2$. All arms train on a fixed subset of our
pretraining data at global batch 256 and are scored with one protocol: EMA weights, 25-step FlowDPM++
sampling, CFG 4.5, GenEval~\cite{geneval}, DPG-Bench~\cite{ella}, FID-10K against a held-out split,
and CLIP score. Each arm is compared with its own control at matched steps.

\paragraph{Prediction target.}
Our first ablation compares $x$- and $v$-prediction. JiT~\cite{jit} argues that predicting the clean
image is necessary once patch tokens are high-dimensional, and Jiang et al.~\cite{empiricalstudy}
find $x$-prediction consistently stronger when converting latent models to pixel space. In our
setting we see no such benefit (Fig.~\ref{fig:prediction}): $v$-prediction trains stably, the two
targets are tied on GenEval at every milestone, and $x$-prediction is worse on DPG and FID.
Following JiT, which trains on noisier timesteps than the common default, the $x$-prediction arm
also uses JiT's timestep distribution: logit-normal $(0.8, 0.8)$ over the noise level instead of
our default $(0.0, 1.0)$, which shifts training toward
noisier timesteps. The contrast therefore does not isolate the prediction target alone. We keep
$v$-prediction.

\paragraph{Representation alignment.}
We then compare four alignment objectives
(Fig.~\ref{fig:ablation}). The control is REPA~\cite{repa} with a frozen DINOv2 teacher. The other
arms keep everything else fixed and change only the alignment: iREPA~\cite{irepa}, which adds a
convolutional projector and spatial normalization; REPA with a DINOv3~\cite{dinov3} teacher; and
Self-Flow~\cite{selfflow}, which replaces the external teacher with a self-supervised objective.

REPA with DINOv2 is the strongest overall. It leads on GenEval at every milestone and is level
with the best arm on DPG and CLIP. Switching the teacher to DINOv3 lowers GenEval at every
milestone and worsens FID throughout; its small DPG lead at the end is within evaluation noise.
iREPA is the one arm with a real gain on any metric: its FID is better than the control's at 150K
and 200K, but it trails on GenEval and CLIP at nearly every milestone, and its FID advantage has
almost vanished by 250K. Self-Flow is far behind on all four metrics. It starts near zero on
GenEval and closes much of the gap as training proceeds, but remains well below the control at
250K. We keep REPA with DINOv2.

\subsection{Scaling to Iris-3B}
\label{sec:arch}
We then scale the selected training recipe ($v$-prediction with DINOv2 REPA) to 3B parameters.
The architectural changes below target efficiency at scale and were not selected by the 256px
ablations. Iris-3B is a
flow-matching~\cite{flowmatching} transformer that operates directly on RGB pixels (Fig.~\ref{fig:arch}).
The image is split into $16{\times}16$ patches, so a $1024^2$ image becomes a $64{\times}64$ token
grid; PixelUMM~\cite{pixelumm} likewise finds that $32{\times}32$ patches train to a higher loss
than $16{\times}16$ even when they see four times as many images. The trunk follows the hybrid
layout of FLUX.1~\cite{flux1}: 8 dual-stream MMDiT
blocks~\cite{sd3}, in which text and image tokens keep separate weights, followed by 16
single-stream blocks, at width 2560. Attention uses GQA~\cite{gqa} with 20
Q and 5 KV heads, a sigmoid output gate~\cite{gatedattn} and sandwich
normalization~\cite{cogview} with RMSNorm~\cite{rmsnorm}, following recent large text-to-image
models~\cite{zimage,krea2}. Timestep modulation replaces the per-block adaLN of DiT~\cite{dit}
with one shared projection per stream (image and text) plus per-block learned biases
(\emph{shared-bias} adaLN), in the spirit of adaLN-single~\cite{pixartalpha} and the light bias
modulation of Krea~2~\cite{krea2}; at equal FLOPs this removes 28\% of the parameters of the same
model with per-block adaLN.

A four-block pixel transformer (PiT) head~\cite{pixeldit} decodes the trunk output to pixels. Each
block modulates every pixel from its patch token and runs a per-pixel MLP; for attention, the
pixels of each patch are compacted into one 1280-wide token, attended across the patch grid and
expanded back. The text encoder is a frozen
Qwen3-VL-4B-Instruct~\cite{qwen3vl}. We read 12 of its layers and fuse them with layerwise
attention pooling~\cite{unifusion} followed by a two-block refiner. Image
tokens use 2D axial RoPE~\cite{rope,ropevit}, which has no parameters tied to resolution, so the
same weights run at every stage of the curriculum.

\end{multicols}
\begin{center}
\includegraphics[width=\linewidth]{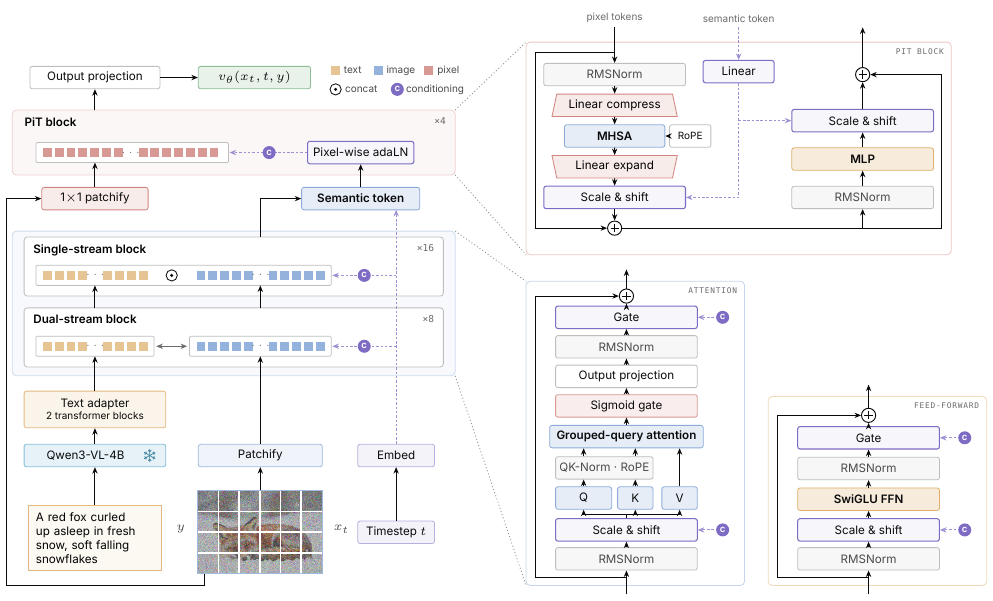}
\captionof{figure}{Iris-3B architecture. \emph{Left:} the full model, with a hybrid dual/single-stream
trunk over $16{\times}16$ pixel patches and a PiT pixel head. \emph{Top right:} the PiT block of
PixelDiT~\cite{pixeldit}. \emph{Bottom right:} a trunk block. Details in Sec.~\ref{sec:arch}.}
\label{fig:arch}
\end{center}
\raggedcolumns
\begin{multicols}{2}

\subsection{Training}
\label{sec:training}

\paragraph{Data.}
We pretrain on a large collection of English image-text pairs that mixes real photographs,
synthetic images from recent generators, and rendered-text images added to strengthen text
rendering. Each stage uses its own set of aspect-ratio buckets. The $256^2$ stage uses centered
square crops. The 512 and 1024 stages assign every image to the nearest of 21 shared aspect-ratio
buckets (1:4 to 4:1, with nearly equal token counts); only images whose short edge reaches the
stage resolution are admitted. Rare
buckets within a source are merged so that every (source, bucket) cell holds at least one global
batch. Because fewer images meet the resolution threshold, non-photographic content dominates the
1024 stage. For supervised fine-tuning (SFT) we use a curated set of high-quality synthetic and real
images whose captions are filtered for quality and prompt alignment.

\paragraph{Objective.}
Given a clean image $x_0$, noise $\epsilon\sim\mathcal{N}(0,I)$ and a caption $c$, we interpolate
along the rectified-flow path
\begin{equation}
x_\sigma=(1-\sigma)\,x_0+\sigma\,\epsilon
\end{equation}
and train the network to predict the velocity $v=\epsilon-x_0$:
\begin{equation}
\mathcal{L}_{\mathrm{flow}}=\mathbb{E}_{x_0,\epsilon,\sigma}\Big[\big\|v_\theta(x_\sigma,\sigma,c)-(\epsilon-x_0)\big\|_2^2\Big]
\end{equation}
The noise level is drawn from a shifted logit-normal distribution:
\begin{equation}
\sigma=\frac{s\,u}{1+(s-1)\,u}\qquad u\sim\text{logit-normal}(0,1)
\end{equation}
with shift $s=2,3,4$ at 256, 512 and 1024. The total loss adds REPA:
\begin{equation}
\mathcal{L}=\mathcal{L}_{\mathrm{flow}}+\lambda\,\mathcal{L}_{\mathrm{REPA}}
\end{equation}
where $\mathcal{L}_{\mathrm{REPA}}$ is the negative cosine similarity between projected trunk tokens
and DINOv2-B/14 patch features, with $\lambda=0.5$ in the 256 stage and $\lambda=0$ afterwards.

\paragraph{Optimization and curriculum.}
We train with hybrid Muon~\cite{muonblog,muon}: Muon (momentum 0.95, Nesterov, RMS-matched LR)
for the hidden weight matrices and AdamW ($\beta_1{=}0.9$, $\beta_2{=}0.95$) for embeddings, norms
and biases, with weight decay 0 and gradient clipping at 0.5. The LR is a constant
$10^{-4}$ after a 2K-step linear warmup. We keep an EMA of the weights with decay 0.9999 and drop the
caption for 10\% of samples to enable CFG. We train in three resolution stages
(Tab.~\ref{tab:stages}). From the 512 stage onward, RoPE uses isotropic
coordinates for non-square images. For the 1024 stage we halve the LR to
$5{\times}10^{-5}$ and reduce the global batch to 512.

\paragraph{SFT.}
We fine-tuned the final pretraining checkpoint on the curated set at 1024 resolution
(Tab.~\ref{tab:stages}), with a constant LR of $4{\times}10^{-5}$.

\begin{table}[H]\centering
\caption{Training stages. Samples = steps $\times$ global batch.}
\label{tab:stages}
\begin{tabular}{@{}lrrr@{}}
\toprule
\thd{Stage} & \thd{Steps} & \thd{Batch} & \thd{Samples}\\
\midrule
256 & 0--367K & 1024 & 375.8M\\
512 multi-AR & 367K--530K & 1024 & 166.9M\\
1024 multi-AR & 530K--625K & 512 & 48.6M\\
SFT & 625K--665K & 512 & 20.5M\\
\bottomrule
\end{tabular}
\end{table}

\subsection{Evaluation}
\label{sec:t2i}
We scored Iris-3B (EMA) with the official, unmodified evaluators of GenEval~\cite{geneval},
DPG-Bench~\cite{ella}, LongText-Bench~\cite{xomni} and OneIG-Bench~\cite{oneig}. Images were
generated at $1024^2$ with CFG 3, 100 sampling steps and 4 samples per prompt, from the official
prompts without rewriting, and every sample was scored (no best-of-$N$). All scores are
for the released checkpoint (SFT step 665K, Tab.~\ref{tab:stages}).

\begin{table}[H]\centering
\caption{Official benchmarks at $1024^2$. References are copied from the respective papers;
FLUX.1-dev from Qwen-Image~\cite{qwenimage} (GenEval, DPG), X-Omni~\cite{xomni} (LongText) and
OneIG-Bench~\cite{oneig} (OneIG, English overall). $^*$Uses prompt rewriting.}
\label{tab:t2i}
\setlength{\tabcolsep}{4pt}
\begin{tabular}{@{}lcccc@{}}
\toprule
\thd{Model} & \thd{GenEval} & \thd{DPG} & \thd{LongText} & \thd{OneIG}\\
\midrule
FLUX.1-dev 12B~\cite{flux1} & 0.66 & 83.8 & 0.607 & 0.434\\
i1 3B$^*$~\cite{i1} & 0.84 & 86.7 & 0.922 & --\\
Z-Image 6B$^*$~\cite{zimage} & 0.84 & 88.1 & 0.935 & 0.546\\
Qwen-Image 20B~\cite{qwenimage} & 0.87 & 88.3 & 0.943 & 0.539\\
\textbf{Iris-3B} & \textbf{0.798} & \textbf{86.52} & \textbf{0.857} & \textbf{0.540}\\
\bottomrule
\end{tabular}
\end{table}

Iris-3B is competitive with strong latent models on the official benchmarks
(Tab.~\ref{tab:t2i}). It clearly surpasses FLUX.1-dev, is level with i1 on DPG, and trails i1,
Z-Image and Qwen-Image on GenEval, although i1 and Z-Image use prompt rewriting. On GenEval it is
strongest on single objects, two objects and colors, and weakest on position and counting.
English long-text rendering is its clearest gap, behind all three, while on OneIG-EN its overall
score matches Qwen-Image and is just below Z-Image. Samples are shown in Fig.~1 and
App.~\ref{app:t2i}.

\section{Latent-to-Pixel Conversion}
\label{sec:l2p}
Our second route to a pixel-space model converts a pretrained latent model, FLUX.2 Klein base
4B~\cite{flux2}.

\subsection{Setup}
We follow the conversion recipe of Jiang et al.~\cite{empiricalstudy}, which builds on
L2P~\cite{l2p} but trains the full model. The VAE is removed and replaced by $16{\times}16$ pixel
patches, which gives each token the same $16{\times}16$-pixel footprint as the parent's latent
tokenization. A new input projection maps pixel patches into the trunk, and a new pixel head
decodes the trunk output. We ablate two heads, DiP~\cite{dip} and a four-block PiT, and use the
latter for the downstream evaluations to match Iris-3B. The model predicts clean images, trained
with a logit-normal timestep distribution at $1024^2$ on a 1:1 mix of real photographs and images
generated by the step-distilled FLUX.2 Klein 4B, a sibling of the parent rather than the parent
itself.

\subsection{Ridge initialization and trunk LR}
Published conversions~\cite{l2p,empiricalstudy} present the transfer as straightforward: replace
the VAE, train briefly, and the converted model approaches its parent; Jiang et al.\ convert
FLUX.2 Klein 9B this way at a global batch of 128. In our small-batch probes on FLUX.2 Klein base
4B it is not. Under the setup above, a 1K-step probe still generates prompt-independent
patch-grid noise, and the unmodified recipe needs more than ten times as many steps to approach
the loss that the recipe below reaches in 1K. We trace this to the input interface. The trunk was
trained on embeddings of VAE latents, so a randomly initialized input projection feeds it inputs
it has never seen, and for FLUX even the best per-patch linear map from pixels explains only 14\%
of the variance of these embeddings.

\begin{figure}[H]
\includegraphics[width=\linewidth]{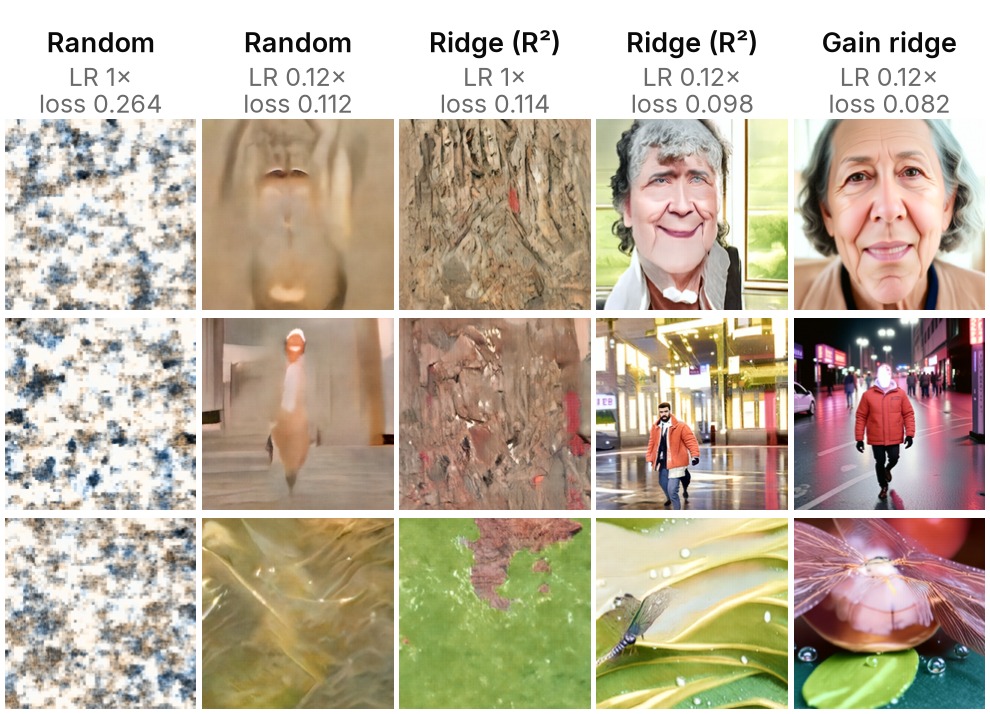}
\caption{FLUX.2 Klein conversion probes after only 1K steps at $1024^2$, one column per input
initialization and trunk LR, with the median loss over steps 900--1000 (PiT head, global batch 8; ``Gain ridge''
is the gain-matched ridge). These are early diagnostic samples used to compare initializations,
not final results; see Fig.~\ref{fig:l2p-final} for the converted model.}
\label{fig:l2p-init}
\end{figure}

Two changes fix this (Fig.~\ref{fig:l2p-init}). The first is a \emph{gain-matched ridge
initialization} of the new input projection. We fit, by ridge regression, a linear map from each
$16{\times}16$ RGB patch to the parent's own embedding of the matching latent token, and choose the
regularization so that the map passes noise with the same gain as the parent's latent path, rather
than to maximize held-out $R^2$, which amplifies pixel noise. Like the Procrustes initialization
of AsymFlow~\cite{asymflow}, this fits the new projection to the parent's latent interface. The
second change scales the trunk LR by the ratio of the parent's weight RMS to that of
Z-Image~\cite{zimage}, our reference latent parent, which for FLUX is $0.12\times$ the head's rate.
FLUX's trunk weights are much smaller, so the same Adam step moves them several times further
relative to their size; the scaled rate equalizes this relative update. On its own, a ridge
initialization fit for held-out $R^2$ more than halves the 1K-step loss, and so does the scaled
LR alone, but neither alone produces coherent images. Together they turn patch noise
into photorealistic, prompt-aligned images within 1K steps, and the gain-matched variant also
removes a colour cast left by the $R^2$-tuned fit. These probes used a global batch of 8 and 1K
steps; we did not test whether the benefit survives at the global batch of 128 used by Jiang et
al.~\cite{empiricalstudy}.

\subsection{Production conversions}
\begin{figure}[H]
\includegraphics[width=\linewidth]{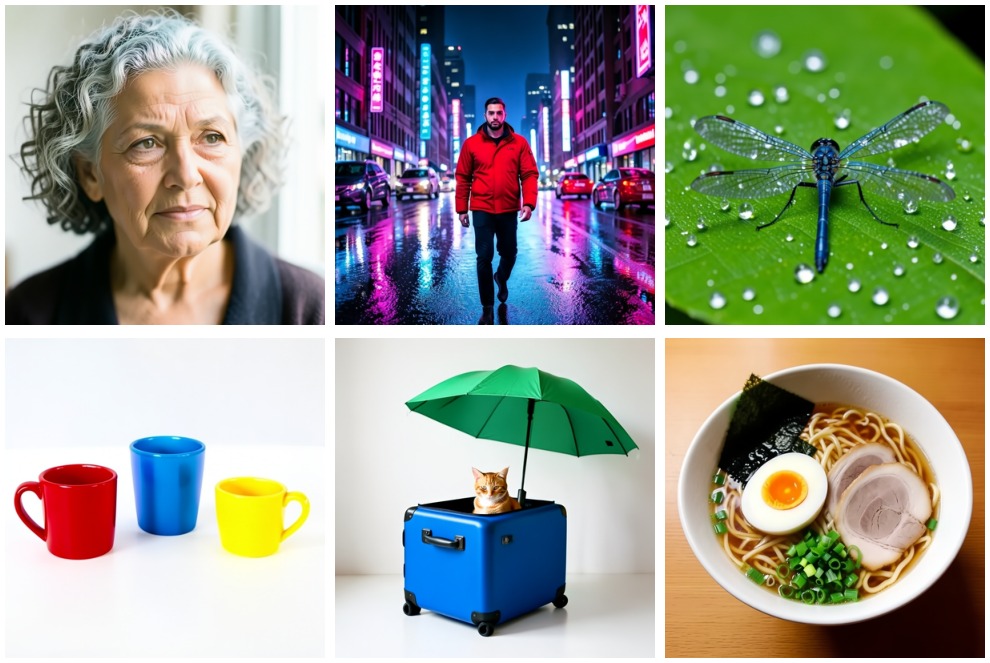}
\caption{Samples from PiT 30K, the converted FLUX.2 Klein used downstream, at $1024^2$ (CFG 4,
100 steps), generated directly in pixel space.}
\label{fig:l2p-final}
\end{figure}

With this recipe we converted FLUX.2 Klein with both heads. Both produce photorealistic images and
render short words correctly. For the production runs we use a global batch of 112, caption
dropout 0.1 and a 1K-step trunk warmup. We use the PiT-head conversion after 30K steps (\emph{PiT 30K}) as the
converted parent for the downstream tasks, matching the head of Iris-3B
(Fig.~\ref{fig:l2p-final}). We have not measured GenEval, DPG or FID for the converted models, so
we cannot say how close they come to their latent parent.

\section{Downstream Tasks}
\label{sec:downstream}
Our hypothesis is that direct pixel generation should beat latent models on tasks where fine
detail matters, since a latent backbone must decode every output through a lossy autoencoder. We
test this by fine-tuning both pixel-space models and comparing them with latent counterparts.

\subsection{Protocol}
We fine-tune two pixel-space parents, Iris-3B and the converted FLUX.2 Klein (PiT 30K),
on monocular depth estimation and image restoration. We choose these two tasks because they are
where a VAE should hurt most. Both are dense, pixel-aligned predictions: depth must place
thin structures and object boundaries at the right pixel, and restoration must synthesize the
high-frequency texture and fine detail that an $8\times$ downsampling autoencoder compresses most
aggressively. Both are established applications of latent generative
priors~\cite{marigoldv2,hypir}. A clean test of the hypothesis needs a latent twin of each parent
trained with the same downstream recipe, data and budget. Tab.~\ref{tab:coverage} shows which
comparisons we have. Two are matched, both on FLUX.2 Klein.

\begin{table}[H]\centering
\caption{Downstream coverage. \checkmark: scored; n/a: no latent twin exists.}
\label{tab:coverage}
\begin{tabular}{@{}lcccc@{}}
\toprule
& \multicolumn{2}{c}{\thd{Depth}} & \multicolumn{2}{c}{\thd{Restoration}}\\
\thd{Parent} & \thd{Pixel} & \thd{Latent} & \thd{Pixel} & \thd{Latent}\\
\midrule
Iris-3B & \checkmark & n/a & \checkmark & n/a\\
FLUX.2 Klein & \checkmark & \checkmark & \checkmark & \checkmark\\
\bottomrule
\end{tabular}
\end{table}

Iris-3B has no latent twin by construction. For depth we train it with the same recipe as the two
FLUX.2 Klein arms, which puts all three on equal footing but does not isolate the representation
space, since the parents differ in size, pretraining data and compute.

\subsection{Monocular depth}
\label{sec:depth}

\begin{figure*}[t]
\includegraphics[width=\linewidth]{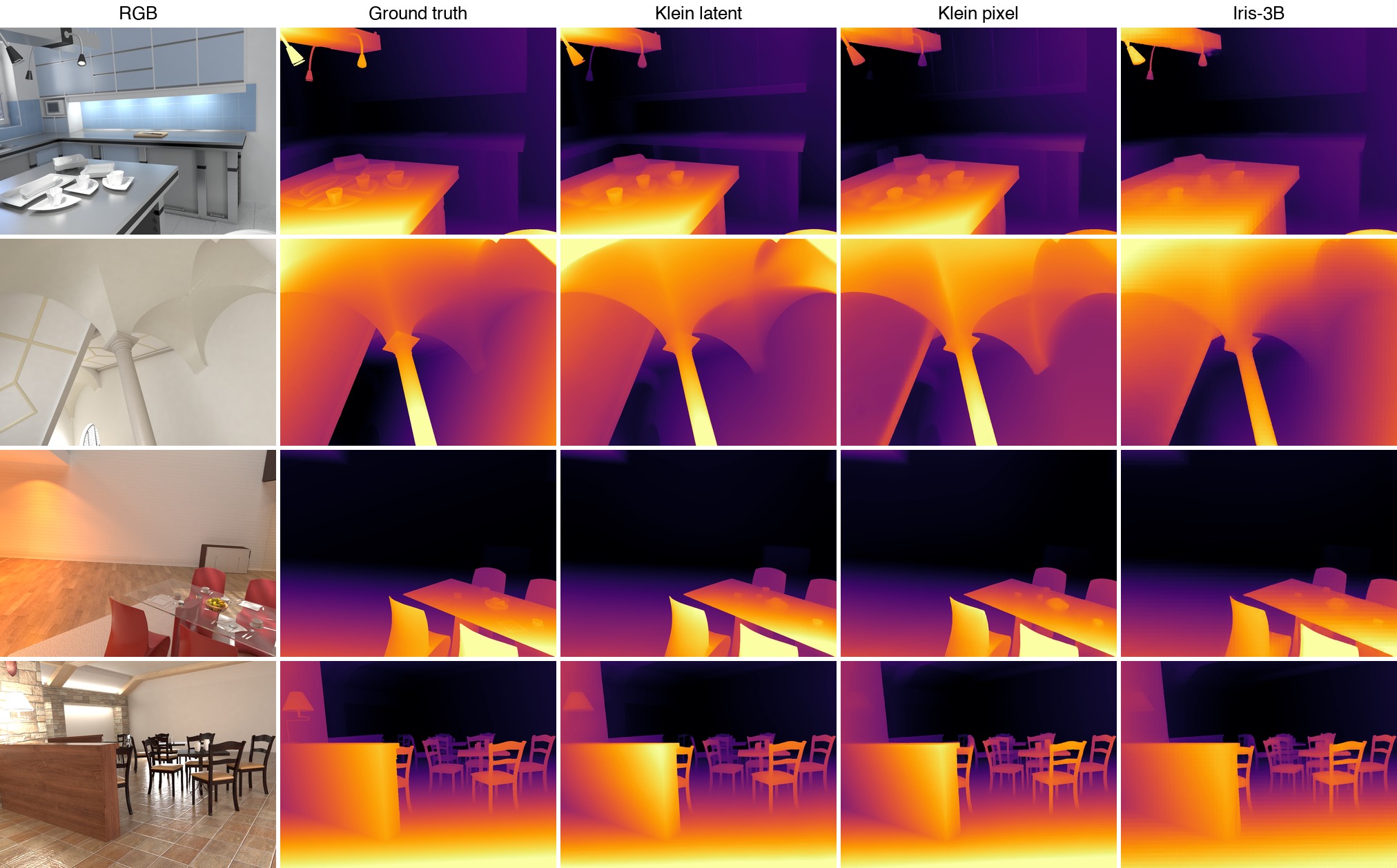}
\caption{Depth on four Hypersim test frames from the three arms of Tab.~\ref{tab:depth}, each
trained with the same direct-regression recipe for 10K steps. Predictions are fitted to the ground
truth with a scale and shift in log depth and shown as inverse depth on a shared per-row colour
scale; black marks invalid ground truth.}
\label{fig:depth-klein}
\end{figure*}

We compare three parents: the latent FLUX.2 Klein base 4B, its pixel conversion (PiT 30K) and
Iris-3B, labelled Klein latent, Klein pixel and Iris-3B in the tables. All three are fine-tuned with
one recipe and one budget, so the only intended difference is the parent.

\paragraph{Recipe.}
Each model predicts depth $\hat d$ in a single deterministic forward pass from the RGB image with the empty
prompt, at the final timestep and with a zero noise input. The target $d$ is relative log depth,
normalized per image by the 2nd and 98th percentiles of the valid ground truth. The pixel models
read depth from their RGB output through a three-to-one convolution; the latent model first decodes
its output latent with the frozen FLUX.2 VAE and reads depth from the decoded image through the
same convolution. The loss is a masked L1 term plus a weighted L1 term on depth gradients, taken
over the set $M$ of valid pixels:
\begin{equation}
\mathcal{L}_{\mathrm{depth}}=\frac{1}{|M|}\sum_{p\in M}\Big(\big|\hat d_p-d_p\big|+5\,\big\|\nabla\hat d_p-\nabla d_p\big\|_1\Big)
\end{equation}
We fully fine-tune with AdamW ($\beta = (0.9, 0.999)$, weight decay 0.01) at a constant
LR of $3{\times}10^{-5}$ without warmup, gradient clipping at 1.0, global batch 32 and
10K steps, and evaluate the raw weights. We chose direct regression after it clearly beat a flow
objective on the converted parent; the choice among L1-family losses made no meaningful
difference.

\paragraph{Data.}
Training uses only synthetic data: Hypersim and Virtual KITTI 2 at $480{\times}640$ and
$352{\times}1216$, mixed 9:1 per batch, with random horizontal flips. Pixels
that are invalid or beyond the far plane (65\,m and 80\,m) are masked from the loss. All arms see
the same images in nearly the same order; only the batch layout across devices, and with it the
pooling of the L1 loss, differs.

\paragraph{Evaluation.}
We score the five standard zero-shot benchmarks, NYUv2, KITTI, ETH3D, ScanNet and DIODE, with the
unmodified evaluation code of Marigold V2~\cite{marigoldv2}, which fits each prediction to the
ground truth with a per-image scale and shift in log depth. We also report its soft edge error
(SEE) on the Hypersim test split, the metric most sensitive to fine detail. All arms share the same
inputs, preprocessing and scorer.

\begin{table*}[t]\centering
\caption{Depth after 10K steps of the same direct-regression recipe, zero-shot on five benchmarks.
AbsRel $\downarrow$ and $\delta_1$ $\uparrow$ per benchmark, their mean over the five benchmarks,
and SEE at kernel 3 on Hypersim $\downarrow$.}
\label{tab:depth}
\setlength{\tabcolsep}{4pt}
\begin{tabular}{@{}lccccccccccccc@{}}
\toprule
& \multicolumn{2}{c}{\thd{NYUv2}} & \multicolumn{2}{c}{\thd{KITTI}} & \multicolumn{2}{c}{\thd{ETH3D}}
& \multicolumn{2}{c}{\thd{ScanNet}} & \multicolumn{2}{c}{\thd{DIODE}} & \multicolumn{2}{c}{\thd{Mean}}
& \thd{Hypersim}\\
\cmidrule(lr){2-3}\cmidrule(lr){4-5}\cmidrule(lr){6-7}\cmidrule(lr){8-9}\cmidrule(lr){10-11}\cmidrule(lr){12-13}\cmidrule(l){14-14}
\thd{Parent} & AbsRel$\downarrow$ & $\delta_1\uparrow$ & AbsRel$\downarrow$ & $\delta_1\uparrow$
& AbsRel$\downarrow$ & $\delta_1\uparrow$ & AbsRel$\downarrow$ & $\delta_1\uparrow$
& AbsRel$\downarrow$ & $\delta_1\uparrow$ & AbsRel$\downarrow$ & $\delta_1\uparrow$ & SEE$\downarrow$\\
\midrule
Klein latent & \best{.050} & \best{.967} & .086 & .921 & \best{.058} & \best{.971} & \best{.060} & \best{.954} & .107 & .924 & .072 & \best{.947} & .528\\
Klein pixel & .064 & .953 & .087 & .919 & .065 & .960 & .075 & .931 & .117 & .906 & .081 & .934 & .526\\
Iris-3B & .057 & .961 & \best{.083} & \best{.927} & .064 & .965 & .069 & .939 & \best{.082} & \best{.935} & \best{.071} & .946 & \best{.496}\\
\bottomrule
\end{tabular}
\end{table*}

Neither pixel-space model clearly improves on the latent one (Tab.~\ref{tab:depth}). Iris-3B is level
with the latent FLUX.2 Klein on the mean of both metrics: it is better on KITTI and DIODE and has
the lowest edge error, and it is worse on NYUv2, ETH3D and ScanNet. The converted pixel model is
behind the latent model on every benchmark except KITTI, where the two are level, and level with it
on SEE. The predictions are also close
qualitatively (Fig.~\ref{fig:depth-klein}; benchmark examples in App.~\ref{app:depth},
Fig.~\ref{fig:depth-benchmarks}). We therefore find no significant improvement from using
a pixel-space generator as the prior for depth. Each arm is a single run at a short budget, and
the per-benchmark differences between Iris-3B and the latent model go both ways and cancel in the
mean, so they should not be read as a win for either.

The converted model's deficit is the one consistent gap. Its parent was adapted from the latent
model with a short latent-to-pixel conversion, whereas the latent arm starts from the fully
pretrained model, so its prior is likely weaker. Iris-3B, a pixel model pretrained from scratch,
shows no such gap, which points to the conversion rather than to pixel space. We have not measured
the converted model's generation quality (Sec.~\ref{sec:l2p}), so this remains a hypothesis.

The raw depth maps of Iris-3B also show a faint patch grid. It is hard to see in
Fig.~\ref{fig:depth-klein}, where predictions are aligned and colour-mapped, so we measure it
directly: the mean curvature of the predicted depth on the pixels next to a $16{\times}16$ patch
border, divided by its mean everywhere else. A model without a grid scores about one.

\begin{table}[H]\centering
\caption{Patch-grid ratio of the raw depth predictions of Tab.~\ref{tab:depth}: mean absolute
second difference of the prediction on pixels adjacent to $16{\times}16$ patch borders over its
mean elsewhere, both image axes, 60 images per benchmark. 1 means no grid $\downarrow$. NYUv2 and
ScanNet are omitted because the latent model, which has no patch grid, already scores well below
one there.}
\label{tab:seam}
\begin{tabular}{@{}lcccc@{}}
\toprule
\thd{Parent} & \thd{Hypersim} & \thd{KITTI} & \thd{ETH3D} & \thd{DIODE}\\
\midrule
Klein latent & \best{0.97} & \best{0.99} & \best{0.99} & \best{0.98}\\
Klein pixel & 1.31 & 1.44 & 1.36 & 1.34\\
Iris-3B & 2.14 & 1.61 & 2.21 & 2.25\\
\bottomrule
\end{tabular}
\end{table}

The latent model has no grid, and both pixel-space models do, Iris-3B most strongly
(Tab.~\ref{tab:seam}). The cause is the patchified design shared by most pixel-space generators.
The input is cut into non-overlapping patches by a pixel-unshuffle, and each patch is decoded
largely from its own trunk token, so neighbouring patches can disagree at their border. The
latent model's VAE decoder, by contrast, has overlapping receptive fields that smooth across
tokens. PixelUMM~\cite{pixelumm} independently reports the same grid in its images and videos,
strongest at high guidance and in smooth regions. It finds that convolutional output heads that
mix neighbouring patches suppress the grid, with PixelShuffle heads giving the best trade-off
against training loss. The artifacts are small next to the depth error itself, but they work
against the fine spatial detail that motivated pixel space in the first place.

\subsection{Restoration and super-resolution}
\label{sec:sr}

\begin{figure*}[t]
\includegraphics[width=\linewidth]{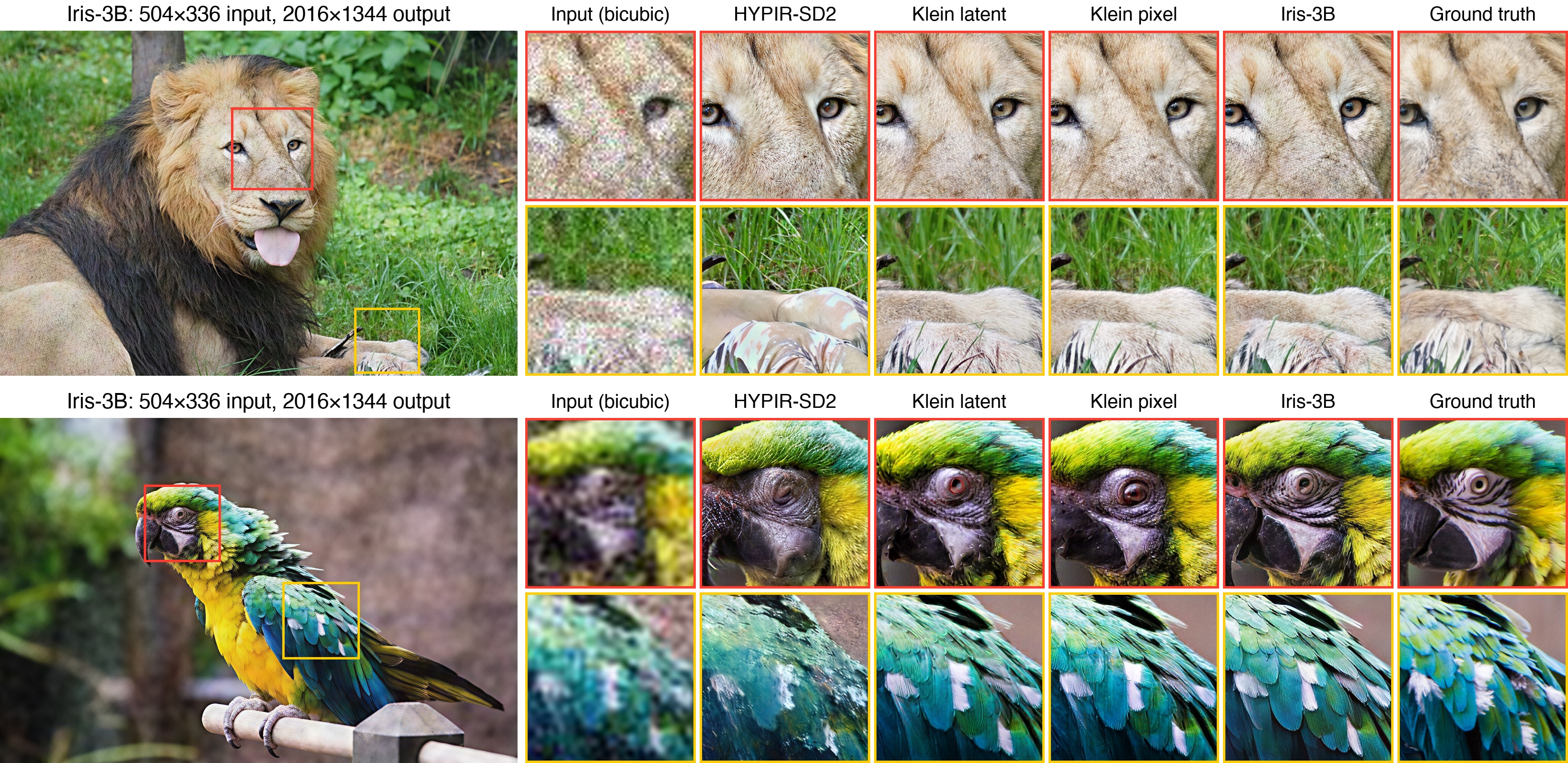}
\caption{$4\times$ restoration of two DIV2K validation images with native
Real-ESRGAN degradations (the Tab.~\ref{tab:sr} pairs). Left: the full Iris-3B output. Right: the
marked regions for each method, cropped from full-resolution outputs. HYPIR-SD2 is the public
release. All methods use an empty
prompt and one forward pass. We chose the images and regions after viewing the outputs.}
\label{fig:sr-div2k}
\end{figure*}

\paragraph{Recipe.}
All restoration models follow the one-step recipe of HYPIR~\cite{hypir}: the degraded image is
upsampled to the target size and a single forward pass with the empty prompt returns the restored
image $\hat y$. Against the clean image $y$, the loss combines a pixel, a perceptual and an
adversarial term:
\begin{equation}
\mathcal{L}_{\mathrm{SR}}=\big\|\hat y-y\big\|_2^2+5\,\mathcal{L}_{\mathrm{LPIPS}}(\hat y,y)+0.5\,\mathcal{L}_{\mathrm{adv}}(\hat y)
\end{equation}
$\mathcal{L}_{\mathrm{LPIPS}}$ uses VGG features, and the critic of $\mathcal{L}_{\mathrm{adv}}$ sits on a frozen OpenCLIP
ConvNeXt-XXL trunk. We fully fine-tune with AdamW at an LR of
$10^{-5}$ for both generator and critic, on $1024^2$ crops with horizontal flips. Training pairs come
from a curated set of high-quality photographs degraded on the fly with the second-order
Real-ESRGAN pipeline~\cite{realesrgan} at $4\times$, including its sharpening.

\paragraph{FLUX.2 Klein, pixel vs.\ latent.}
This is our second matched comparison. We fine-tuned the latent FLUX.2 Klein and its converted
pixel counterpart (PiT 30K) with the same data, losses and schedule: global batch 8 and 500 warmup
steps, 10K steps with a linear decay to zero over the last 2K, then continued to 30K with the
LR reset and decayed again over the last 2K. We score the raw weights. We
evaluate both on DIV2K~\cite{div2k} validation images with native $4\times$ degradations, using
fidelity metrics (PSNR, SSIM, LPIPS) and no-reference quality metrics (NIQE, MUSIQ, DeQA).

The pixel model is not better. It is worse on most metrics in Tab.~\ref{tab:sr}, most clearly SSIM
and NIQE, and level on LPIPS. Each arm is a single run and we ran no significance test, so the
small differences should be read as ``no advantage'' rather than as a reliable loss. The latent arm
is itself a fine-tune rather than the untouched parent.

\paragraph{Iris-3B.}
We also adapted Iris-3B into a one-step restoration model with the same recipe, at a constant
LR for 10K steps, with global batch 40 for the first 5K and 10 thereafter, and score
an EMA of the weights (decay 0.999). On the same DIV2K pairs it is slightly
below both FLUX.2 Klein models on fidelity and DeQA, between them on NIQE, and the best of all
methods in Tab.~\ref{tab:sr} on MUSIQ. Its parent checkpoint, batch, schedule and budget differ
from the FLUX.2 Klein arms, and its inference adds wavelet colour correction and tiling, so this
comparison is not matched.

\begin{table}[H]\centering
\caption{DIV2K $4\times$ restoration on validation images with native $4\times$ Real-ESRGAN
degradations; every method receives the same saved input pairs and is scored with the same PyIQA
metric settings and DeQA-Score revision. PSNR and SSIM are computed on RGB. Public methods run at
their released settings, except that all text conditioning (including tag and caption generators)
is replaced by an empty prompt. Grouped as ours, GAN, multi-step diffusion, one-step diffusion.}
\label{tab:sr}
\setlength{\tabcolsep}{2.6pt}
\begin{tabular}{@{}lcccccc@{}}
\toprule
\thd{Model} & \thd{PSNR$\uparrow$} & \thd{SSIM$\uparrow$} & \thd{LPIPS$\downarrow$} & \thd{NIQE$\downarrow$} & \thd{MUSIQ$\uparrow$} & \thd{DeQA$\uparrow$}\\
\midrule
Klein latent & 20.99 & 0.557 & 0.275 & 3.26 & 67.11 & \best{4.06}\\
Klein pixel & 20.87 & 0.504 & \best{0.274} & 4.11 & 67.03 & 4.04\\
Iris-3B & 20.68 & 0.481 & 0.292 & 3.94 & \best{68.01} & 3.87\\
\midrule
R-ESRGAN~\cite{realesrgan} & 21.76 & 0.579 & 0.386 & 3.79 & 60.37 & 3.85\\
\midrule
StableSR~\cite{stablesr} & 21.91 & 0.557 & 0.407 & 4.10 & 53.91 & 3.59\\
DiffBIR v2~\cite{diffbir} & 21.29 & 0.476 & 0.471 & 3.39 & 66.21 & 3.88\\
SeeSR~\cite{seesr} & \best{22.25} & \best{0.580} & 0.373 & 4.71 & 59.46 & 3.78\\
SUPIR-v0Q~\cite{supir} & 21.53 & 0.501 & 0.386 & 3.91 & 59.64 & 3.79\\
InvSR~\cite{invsr} & 20.18 & 0.534 & 0.426 & 4.43 & 60.94 & 3.57\\
\midrule
OSEDiff~\cite{osediff} & 21.40 & 0.542 & 0.352 & 3.49 & 67.75 & 3.91\\
S3Diff~\cite{s3diff} & 21.77 & 0.551 & 0.311 & 3.45 & 66.70 & 3.89\\
HYPIR-SD2~\cite{hypir} & 19.96 & 0.521 & 0.353 & \best{3.12} & 67.53 & 3.96\\
\bottomrule
\end{tabular}
\end{table}

Overall, the pixel-space models we adapt are not better than latent models on the benchmark
numbers, but they are competitive with open-source restoration methods. Our three models,
HYPIR-SD2 and OSEDiff sit within a narrow band on the no-reference metrics MUSIQ and DeQA. SeeSR
and R-ESRGAN lead on SSIM and are among the best on PSNR, but trail our models and HYPIR-SD2 on
LPIPS and MUSIQ. We show qualitative results on DIV2K in Fig.~\ref{fig:sr-div2k} and, in
App.~\ref{app:sr-real}, on further DIV2K images (Figs.~\ref{fig:sr-div2k-extra1}
and~\ref{fig:sr-div2k-extra2}) and on real-world inputs (Figs.~\ref{fig:sr-reallq}
and~\ref{fig:sr-real}).

\section{Conclusion}
\label{sec:conclusion}
We pretrained Iris-3B, a 3B-parameter pixel-space text-to-image model, from scratch to $1024^2$.
It is competitive with strong latent models on the
official benchmarks, showing that end-to-end pixel-space pretraining with a PiT head scales to 3B
parameters without a VAE. We also converted FLUX.2 Klein to pixel space, which gives a pixel model
and a latent twin from the same parent and makes a controlled comparison possible. Fine-tuned for
image restoration, the converted model trails its latent twin slightly. Fine-tuned for monocular
depth with one matched recipe, neither pixel model improves on the latent FLUX.2 Klein: Iris-3B is
level with it, and the converted model trails it, possibly because the short conversion leaves it
with a weaker prior. We find no significant improvement from a pixel-space generative prior. We
release the weights and training code of Iris-3B as a strong starting point for further work on
pixel-space generation.

\paragraph{Limitations.}
Our downstream comparisons are confounded. Iris-3B has no latent twin, so its depth comparison
with the latent FLUX.2 Klein mixes the representation space with differences in size, pretraining
data and compute. The two matched comparisons share one converted parent, whose generative prior
is likely weaker than that of the latent model it was converted from, so they compare a briefly
converted model with a fully pretrained one. Each depth and restoration arm is a single run at a
short budget with no significance test, and the Iris-3B restoration arm is not matched to the
FLUX.2 Klein arms. Iris-3B was trained with modest compute and data and is not state of the art in terms of
raw generation quality. Finally,
like most pixel-space methods, our models decode the pixels of each patch largely on their own,
which leaves faint patch artifacts in dense predictions.

\paragraph{Future work.}
Three directions follow. First, improving Iris-3B itself, whose quality and speed can be raised
with reinforcement learning and few-step distillation. Second, removing the patch grid. The PiT
head of Iris-3B exchanges information between patches only through one compacted token per patch,
so neighbouring pixels across a patch border are never processed together at pixel resolution, and the grid this
leaves is most visible in dense predictions. An output head with overlapping receptive fields, such as the
convolutional and PixelShuffle heads studied in PixelUMM~\cite{pixelumm}, is the natural fix, and
depth estimation is a direct test of whether it helps. Third, editing and reference-based
generation without a VAE. Editing must reproduce the unchanged parts of a source image exactly,
which is where a lossy latent should hurt most. PixelUMM conditions on clean reference images
given as raw pixel patches, without a separate VAE stream, and Iris-3B already reads its prompt
through a vision-language encoder. Adding editing data and reference images to its training would
give a fully pixel-space editing model. Paired with a latent model pretrained on the same data
with matched compute, it would remove the confounds of this study and test whether pixel space
helps where its advantage should be largest.

\end{multicols}

\clearpage
\appendix
\iflatexml\else\titleformat{\section}{\brock\fontsize{15.5}{18}\selectfont\color{ink}}{\Alph{section}\;·\;}{0pt}{}\fi
\section{Text-to-Image Samples}
\label{app:t2i}
Samples from the final Iris-3B checkpoint (EMA weights after SFT), generated
directly in pixel space at native aspect ratios of about one megapixel with CFG 3 and 100 sampling
steps. Each image uses one fixed seed, was picked from a single rendering of a broad prompt set, and
is shown with its full prompt, without prompt rewriting.

\vspace{3mm}
\noindent \begin{minipage}[t]{0.4900\linewidth}\includegraphics[width=\linewidth]{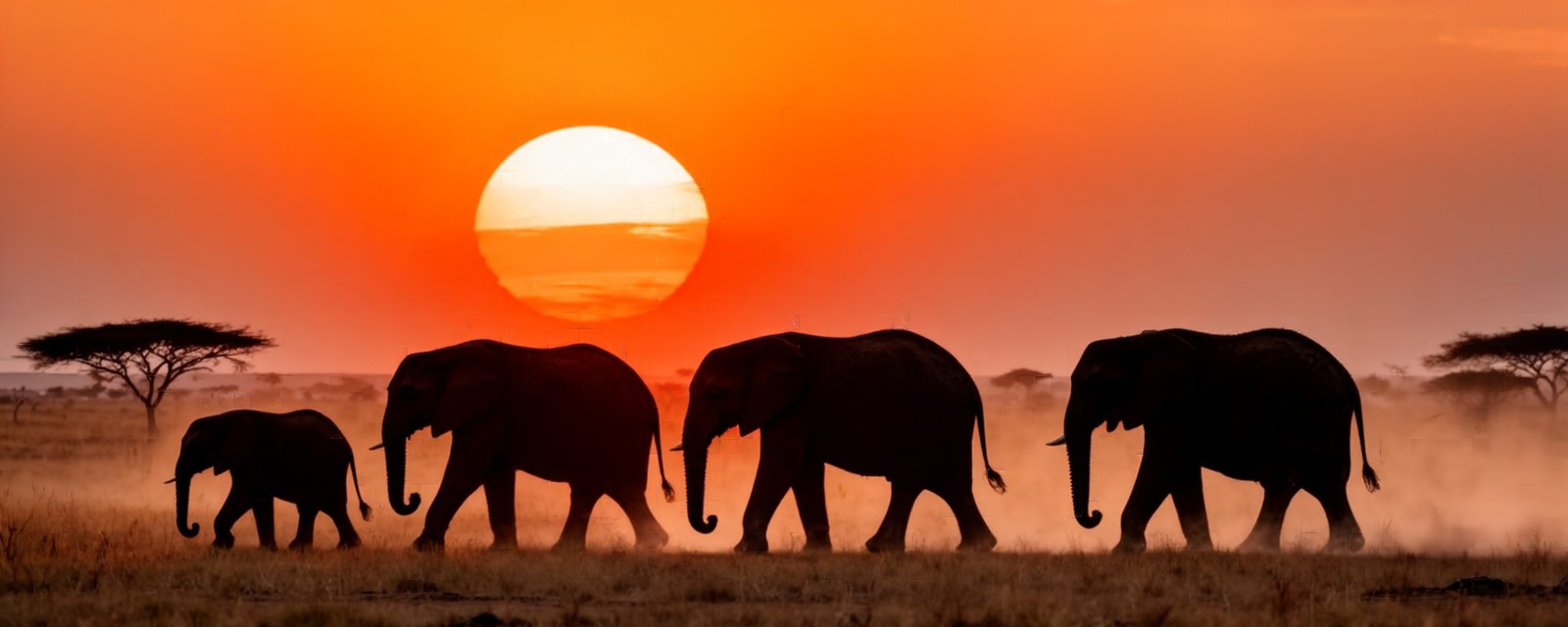}\\[1mm]{\raggedright\scriptsize\color{g600} A family of elephants walking across the African savanna at sunset, silhouettes against a huge orange sun, dust in the air\par}\end{minipage}\hspace{0.02\linewidth}\begin{minipage}[t]{0.4900\linewidth}\includegraphics[width=\linewidth]{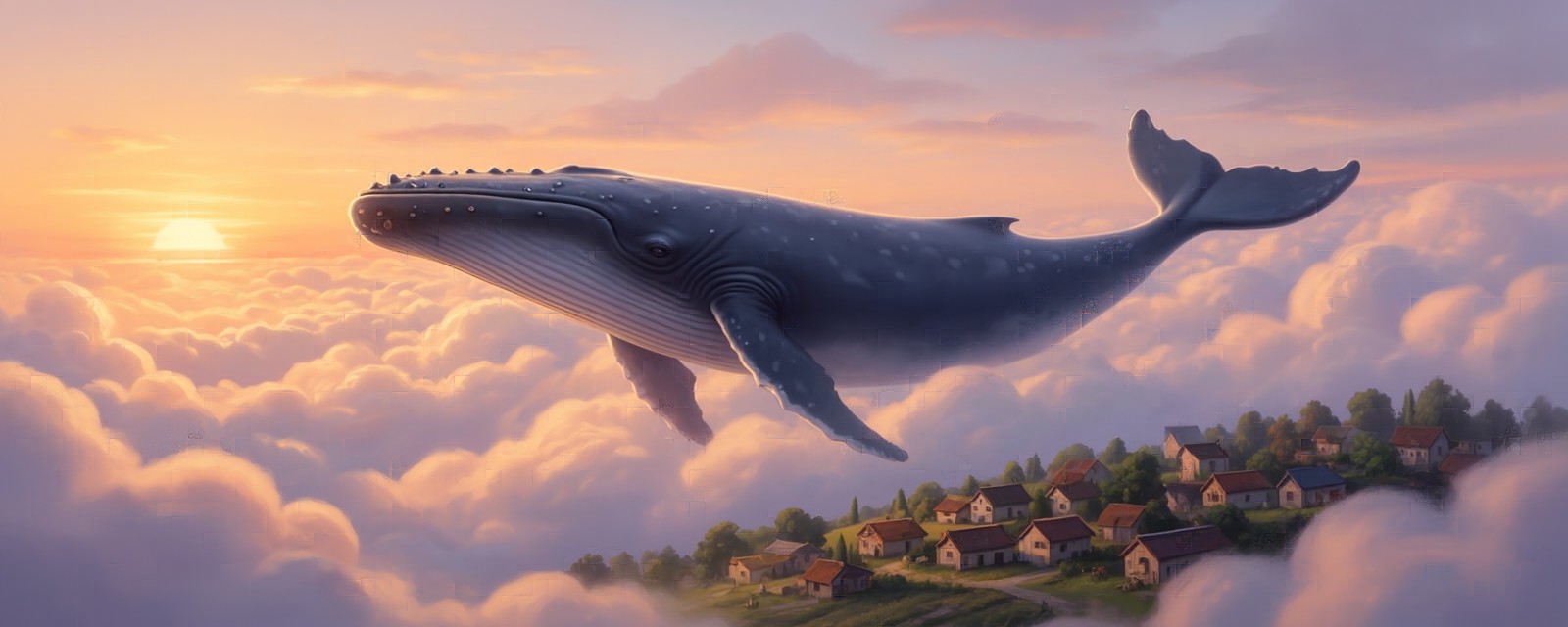}\\[1mm]{\raggedright\scriptsize\color{g600} A giant whale swimming through the clouds above a small village at sunset, surreal fantasy art, soft warm colors\par}\end{minipage}\par
\vspace{4mm}
\noindent \begin{minipage}[t]{0.5337\linewidth}\includegraphics[width=\linewidth]{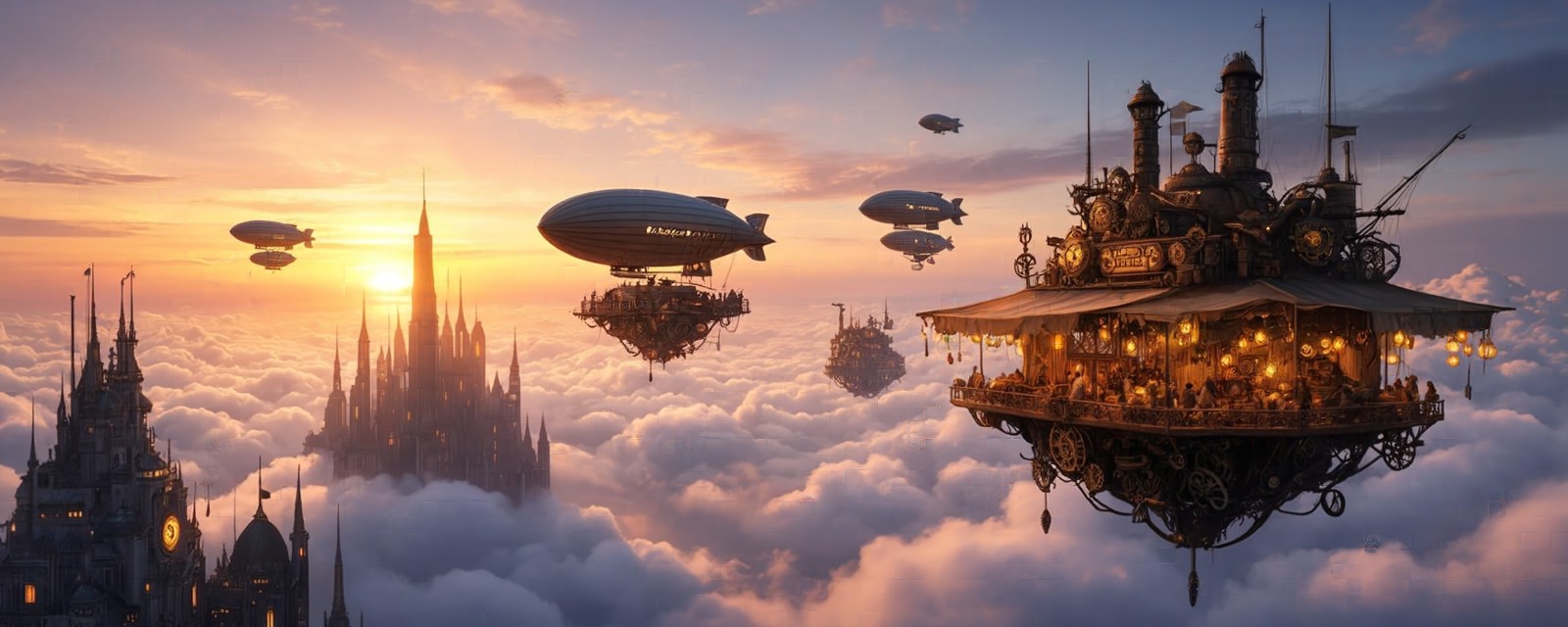}\\[1mm]{\raggedright\scriptsize\color{g600} A floating market of airships above a cloud city at sunrise, steampunk fantasy\par}\end{minipage}\hspace{0.02\linewidth}\begin{minipage}[t]{0.4463\linewidth}\includegraphics[width=\linewidth]{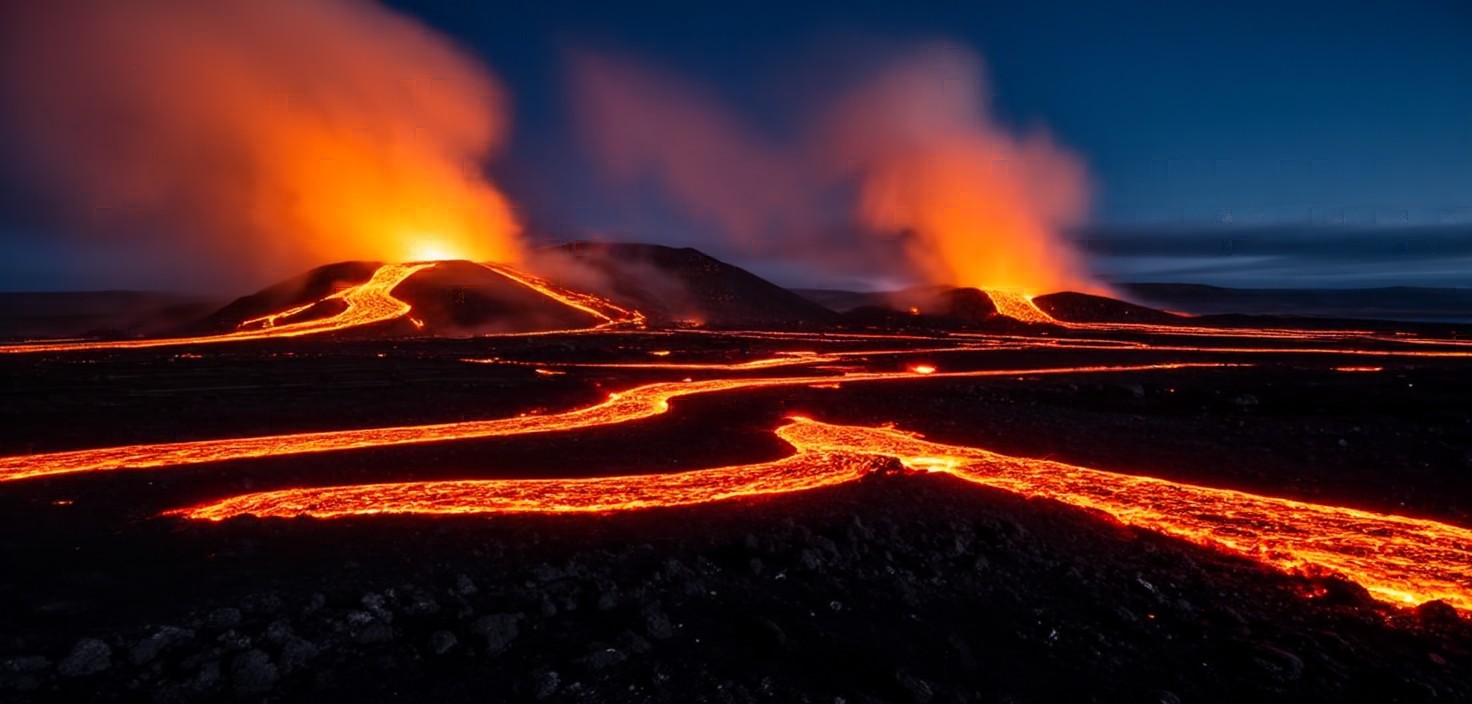}\\[1mm]{\raggedright\scriptsize\color{g600} Volcanic eruption at night in Iceland, rivers of glowing lava flowing across black fields, plumes of steam lit orange, long exposure\par}\end{minipage}\par
\vspace{4mm}
\noindent \begin{minipage}[t]{0.4900\linewidth}\includegraphics[width=\linewidth]{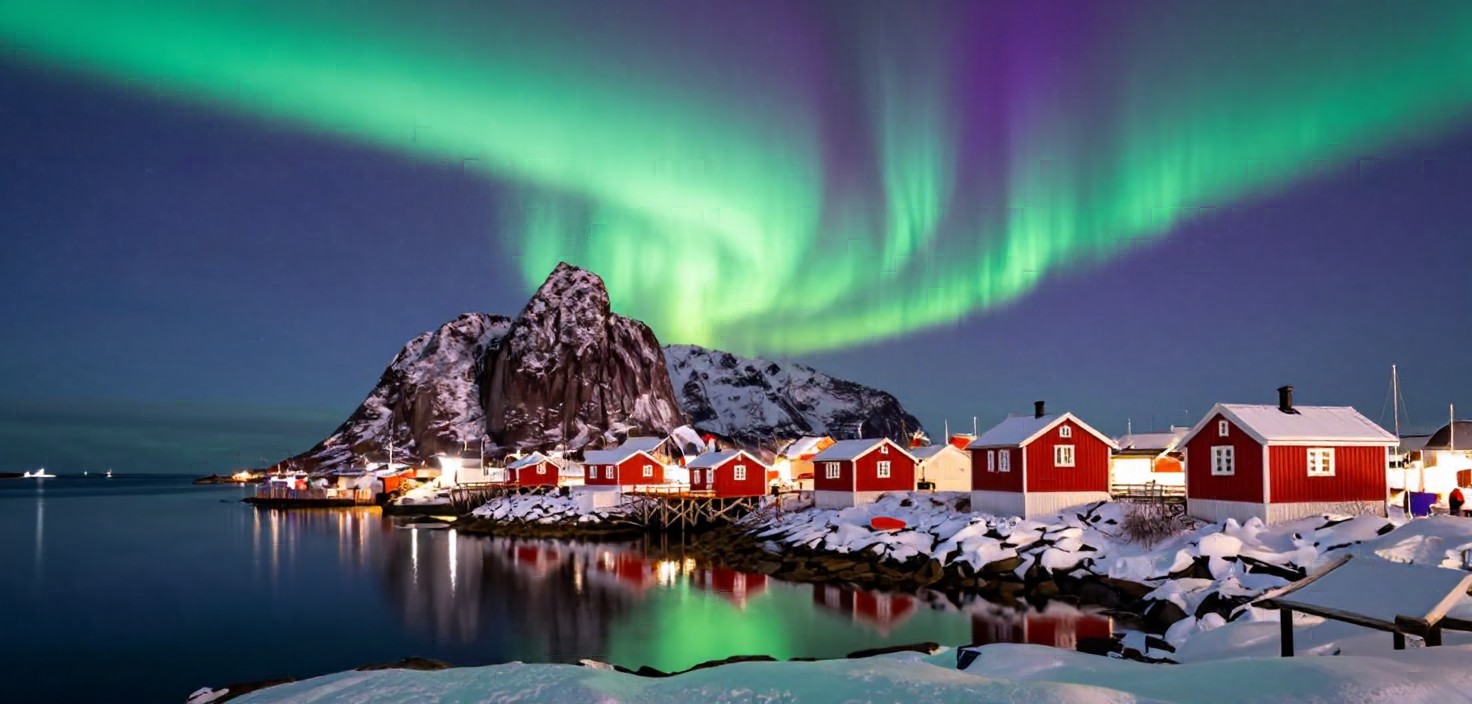}\\[1mm]{\raggedright\scriptsize\color{g600} The aurora borealis swirling green and violet over a snowy Lofoten fishing village with red wooden cabins, reflections in a calm fjord, night photograph\par}\end{minipage}\hspace{0.02\linewidth}\begin{minipage}[t]{0.4900\linewidth}\includegraphics[width=\linewidth]{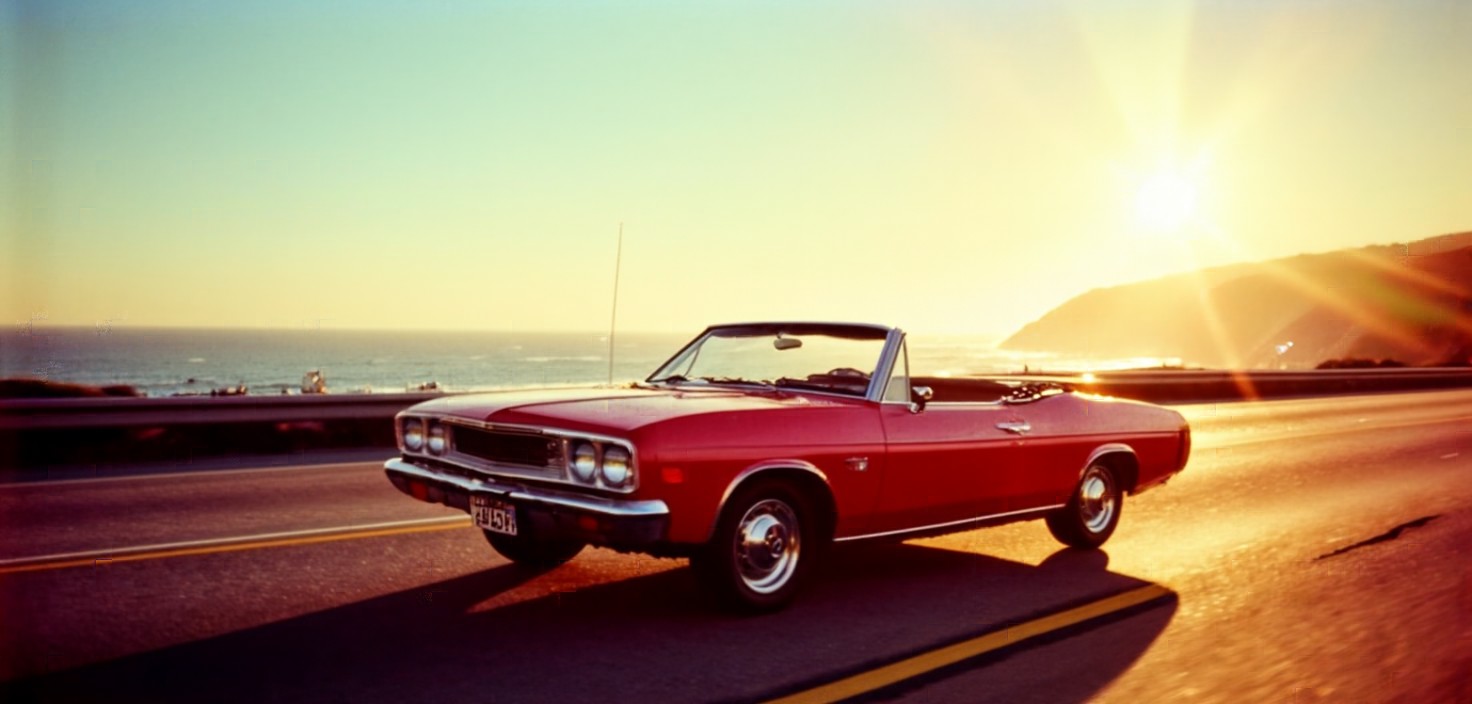}\\[1mm]{\raggedright\scriptsize\color{g600} A vintage red convertible driving along a coastal highway at golden hour, sun flare, 1970s film photograph aesthetic, Kodak Portra colors\par}\end{minipage}\par
\vspace{4mm}
\noindent \begin{minipage}[t]{0.5335\linewidth}\includegraphics[width=\linewidth]{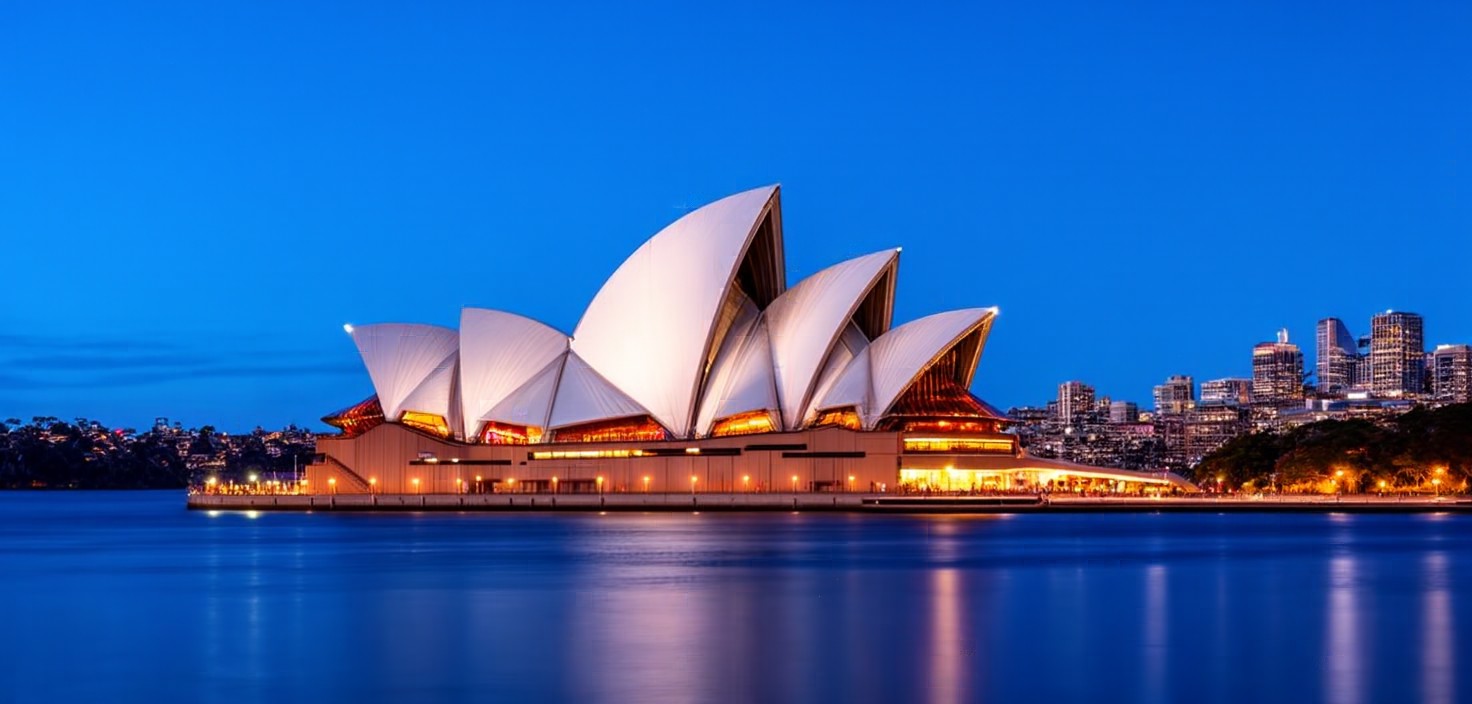}\\[1mm]{\raggedright\scriptsize\color{g600} The Sydney Opera House at blue hour reflecting in the harbor, city lights\par}\end{minipage}\hspace{0.02\linewidth}\begin{minipage}[t]{0.4465\linewidth}\includegraphics[width=\linewidth]{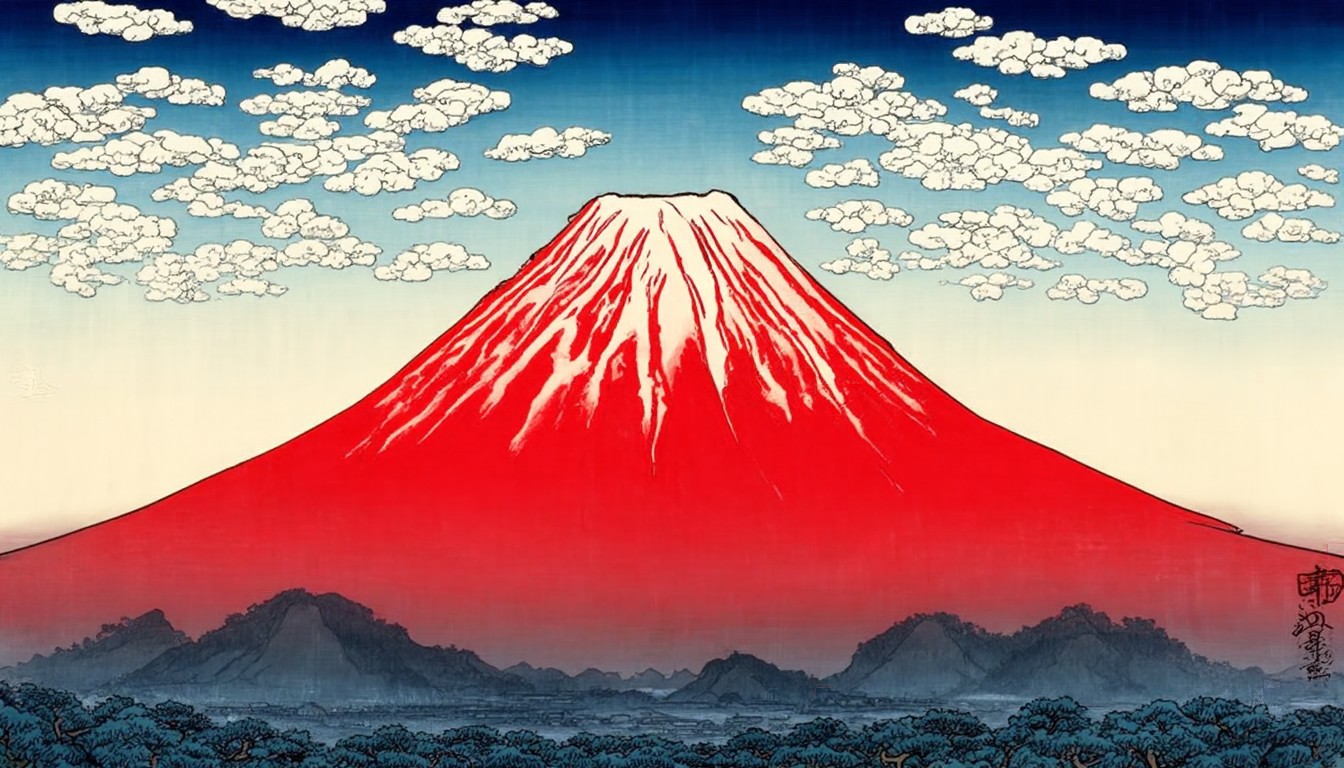}\\[1mm]{\raggedright\scriptsize\color{g600} A Hokusai inspired painting of a red Mount Fuji under a sky of small white clouds\par}\end{minipage}\par
\clearpage
\noindent \begin{minipage}[t]{0.4900\linewidth}\includegraphics[width=\linewidth]{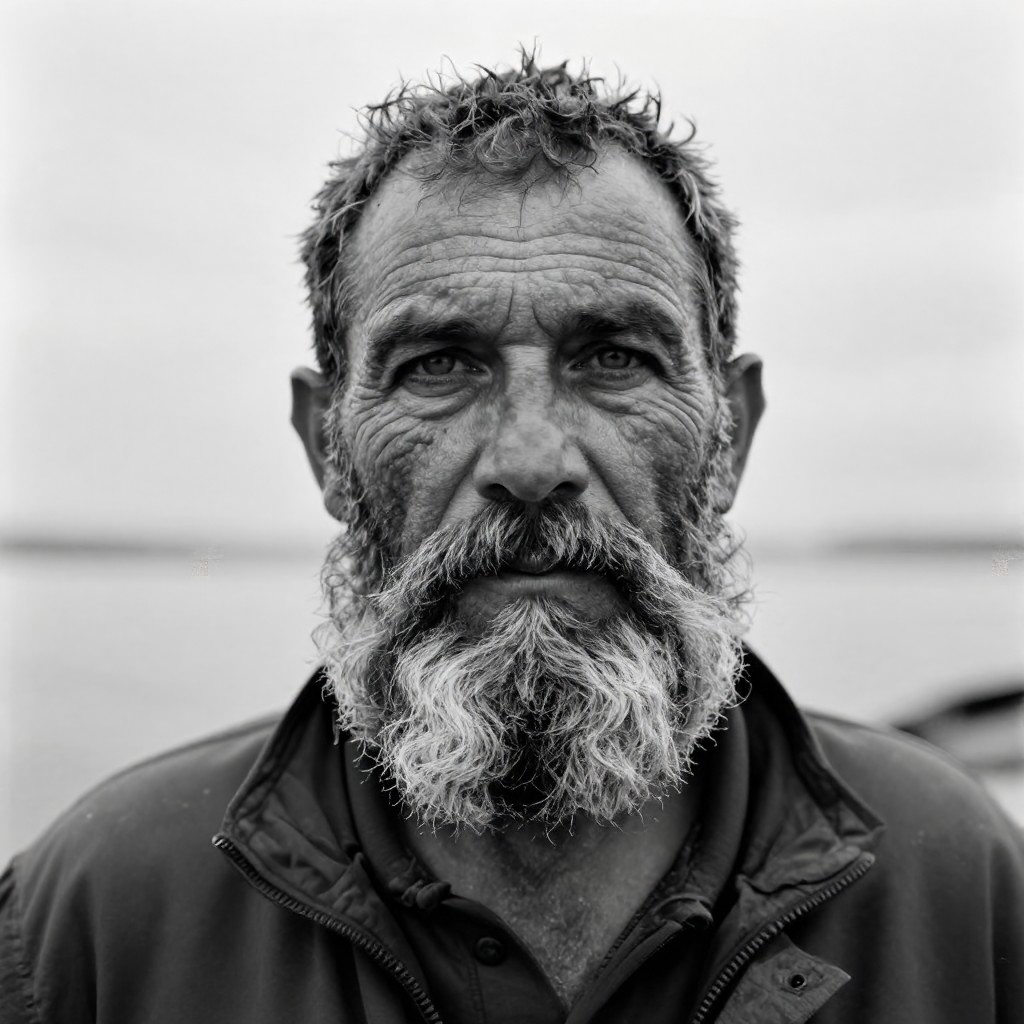}\\[1mm]{\raggedright\scriptsize\color{g600} Black and white portrait of a fisherman with a thick grey beard and deep wrinkles, piercing eyes, overcast light, fine grain film photograph\par}\end{minipage}\hspace{0.02\linewidth}\begin{minipage}[t]{0.4900\linewidth}\includegraphics[width=\linewidth]{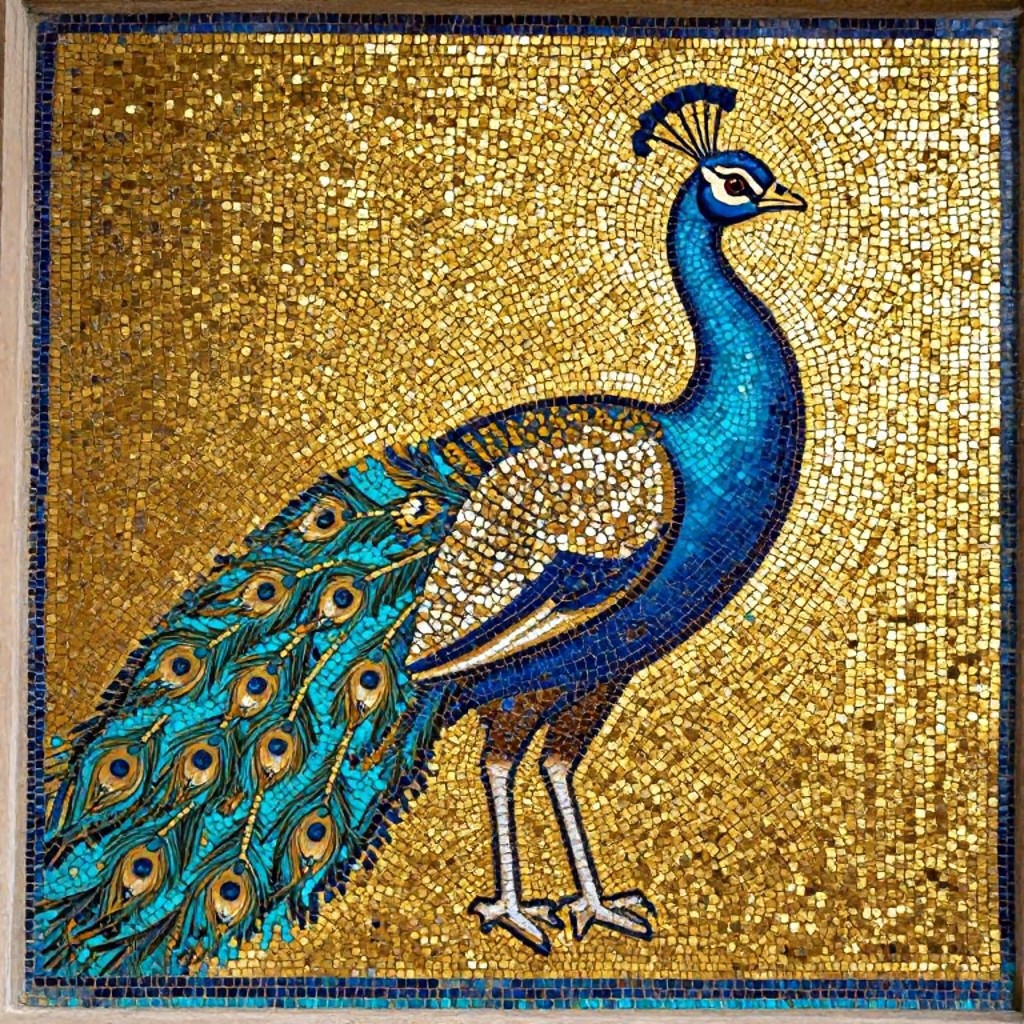}\\[1mm]{\raggedright\scriptsize\color{g600} A Byzantine-style mosaic of a peacock made of tiny gold and turquoise tiles, shimmering texture\par}\end{minipage}\par
\vspace{4mm}
\noindent \begin{minipage}[t]{0.4900\linewidth}\includegraphics[width=\linewidth]{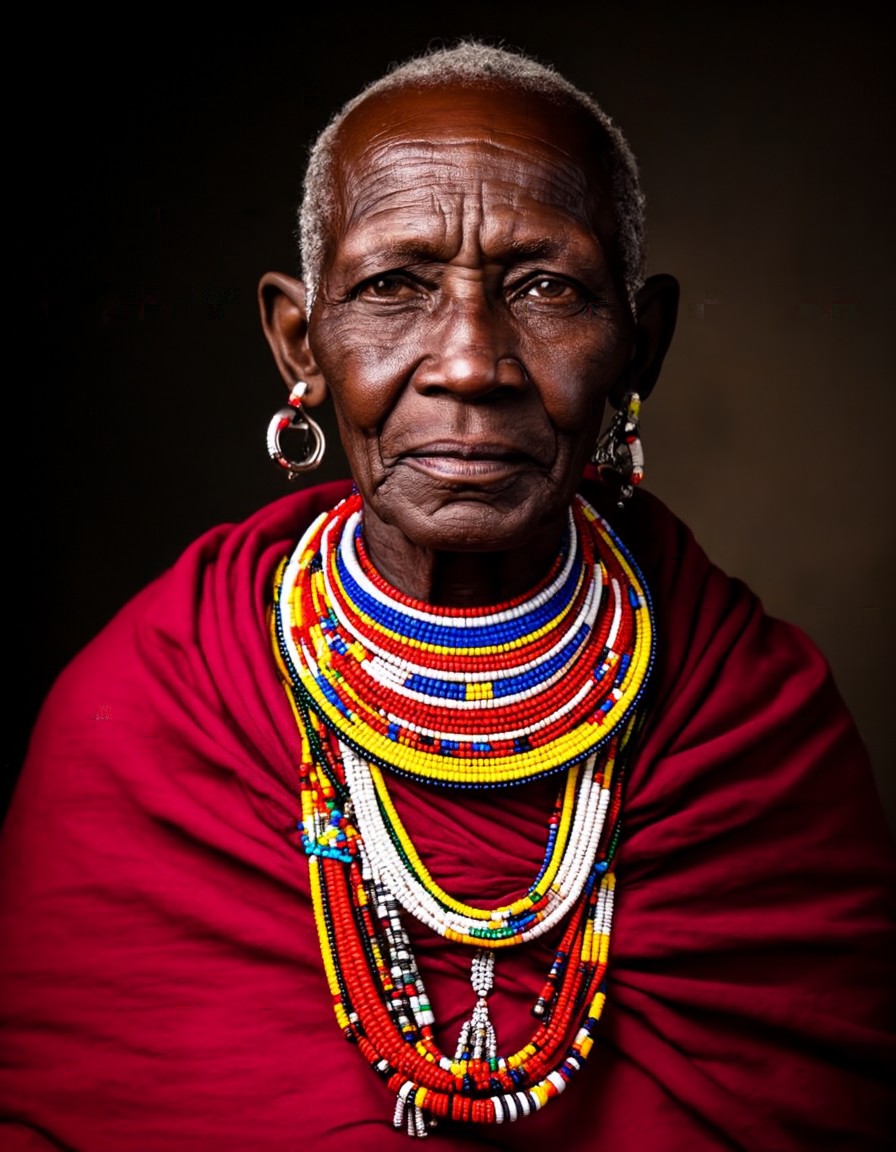}\\[1mm]{\raggedright\scriptsize\color{g600} Studio portrait of a Maasai elder wearing vibrant beaded jewelry, deep red cloth, dark backdrop, Rembrandt lighting, ultra detailed skin texture\par}\end{minipage}\hspace{0.02\linewidth}\begin{minipage}[t]{0.4900\linewidth}\includegraphics[width=\linewidth]{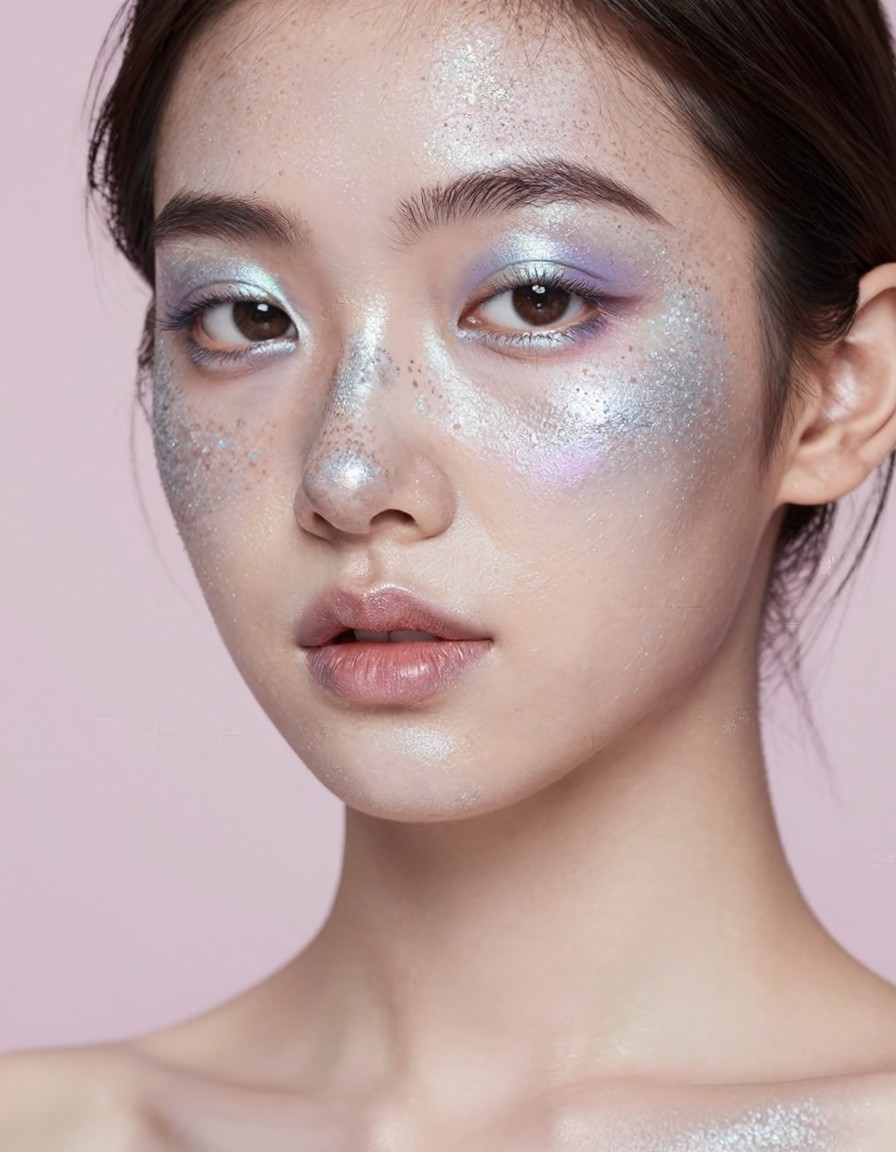}\\[1mm]{\raggedright\scriptsize\color{g600} Close-up portrait of a young woman with silver glitter freckles and iridescent makeup, soft pastel background, high fashion beauty photography\par}\end{minipage}\par
\clearpage
\noindent \begin{minipage}[t]{0.3200\linewidth}\includegraphics[width=\linewidth]{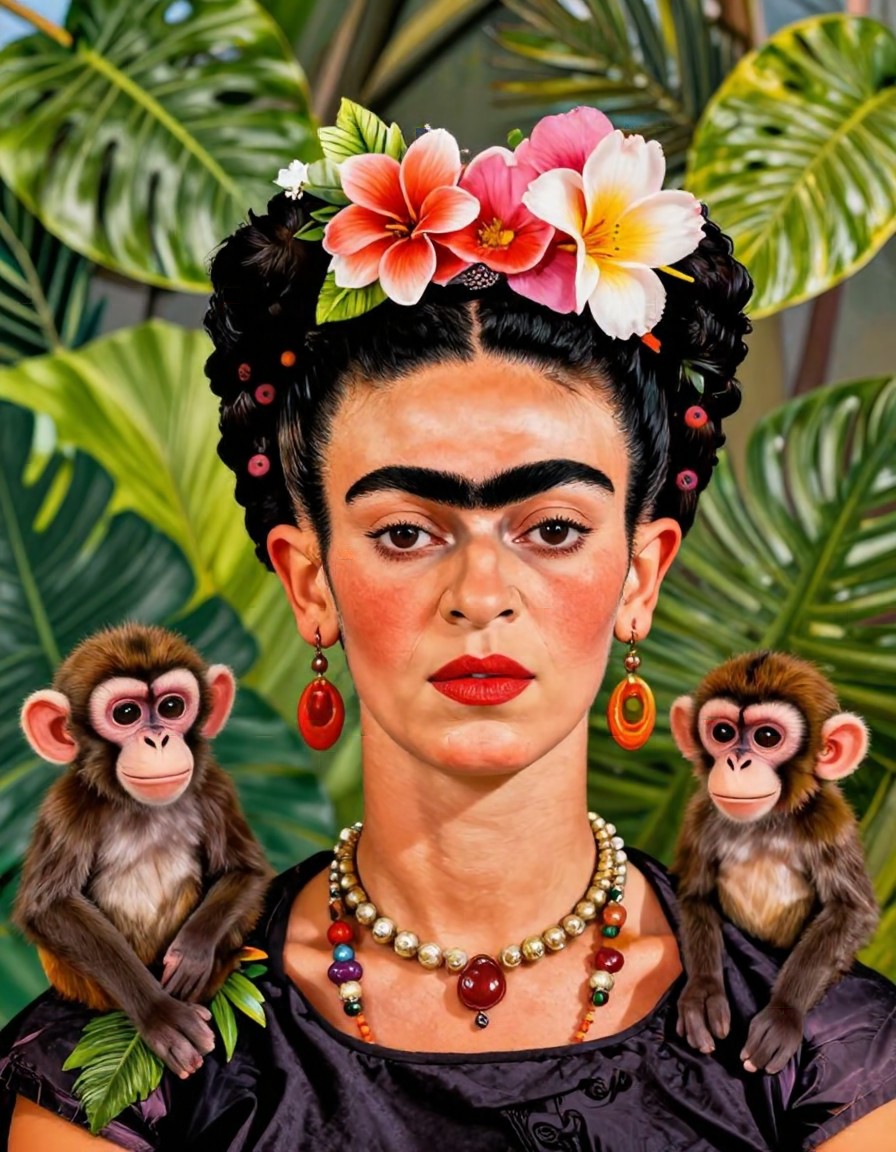}\\[1mm]{\raggedright\scriptsize\color{g600} A Frida Kahlo inspired self portrait with monkeys, tropical leaves and flowers in the hair\par}\end{minipage}\hspace{0.02\linewidth}\begin{minipage}[t]{0.3200\linewidth}\includegraphics[width=\linewidth]{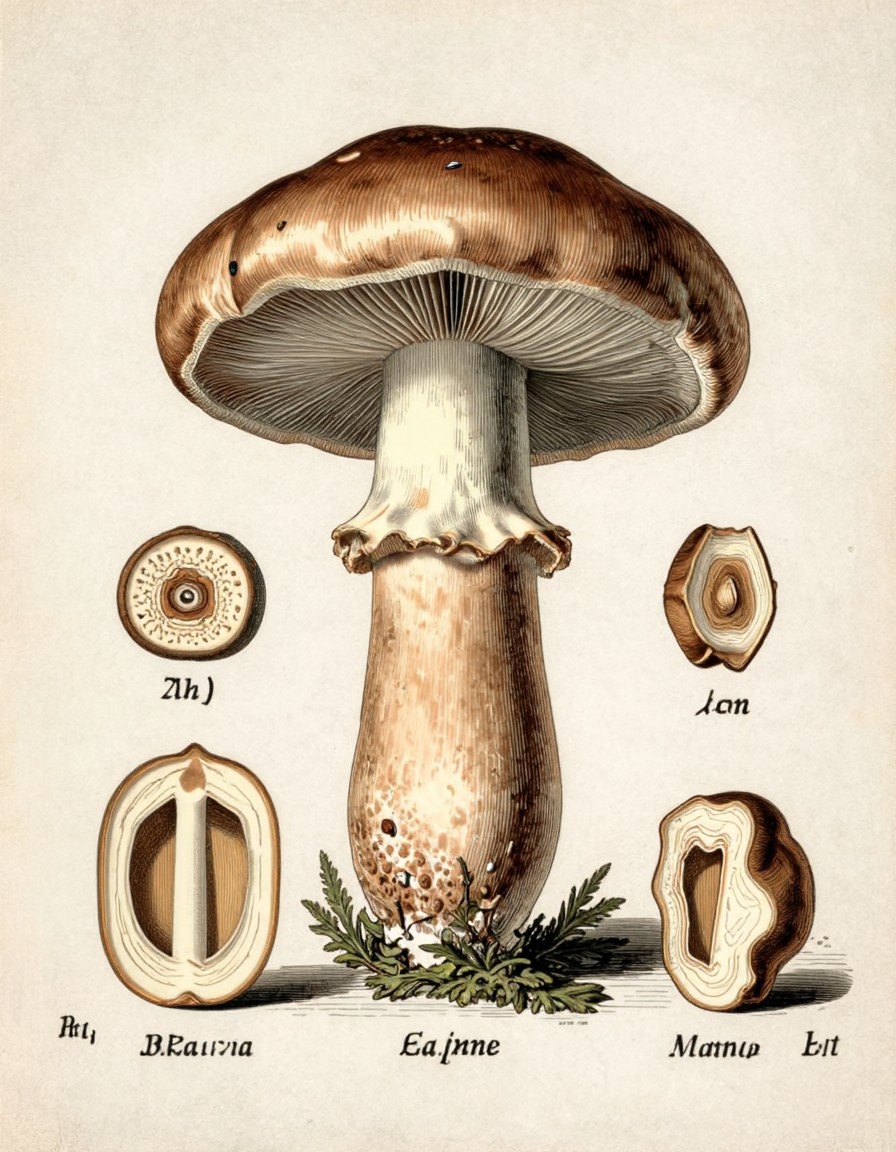}\\[1mm]{\raggedright\scriptsize\color{g600} A botanical scientific illustration of a mushroom with labeled cross-sections, vintage engraving style\par}\end{minipage}\hspace{0.02\linewidth}\begin{minipage}[t]{0.3200\linewidth}\includegraphics[width=\linewidth]{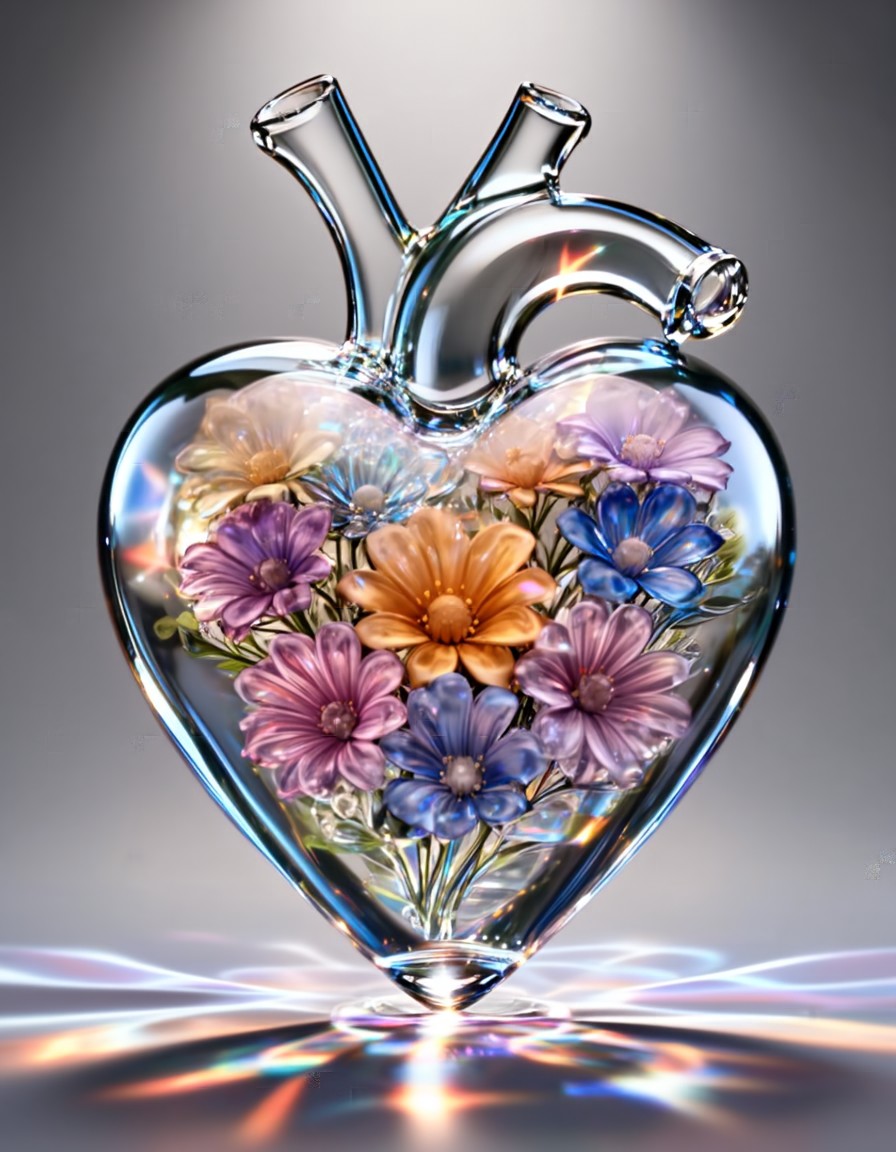}\\[1mm]{\raggedright\scriptsize\color{g600} A glass sculpture of a heart filled with flowers, caustics and reflections, 3D render\par}\end{minipage}\par
\vspace{4mm}
\noindent \begin{minipage}[t]{0.3200\linewidth}\includegraphics[width=\linewidth]{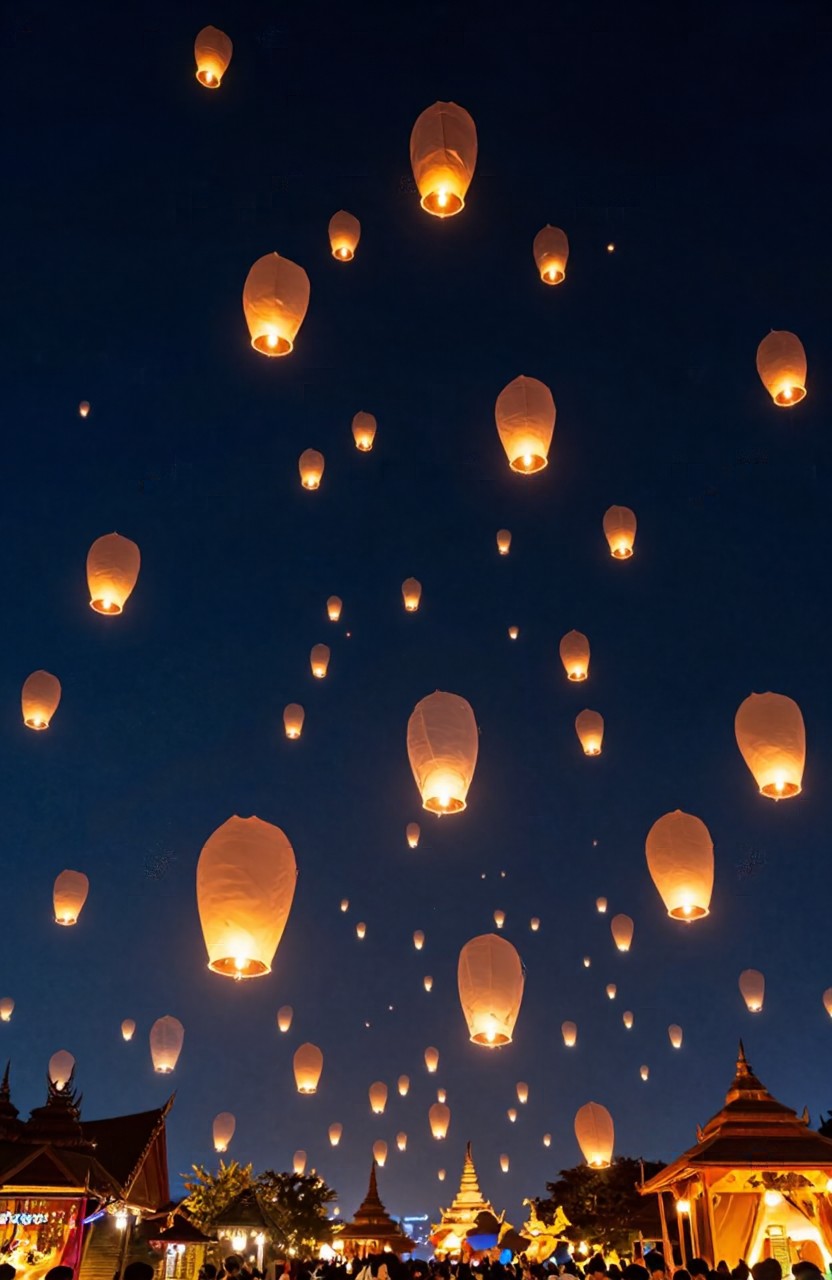}\\[1mm]{\raggedright\scriptsize\color{g600} Thousands of sky lanterns rising into the night sky at Yi Peng festival in Chiang Mai\par}\end{minipage}\hspace{0.02\linewidth}\begin{minipage}[t]{0.3200\linewidth}\includegraphics[width=\linewidth]{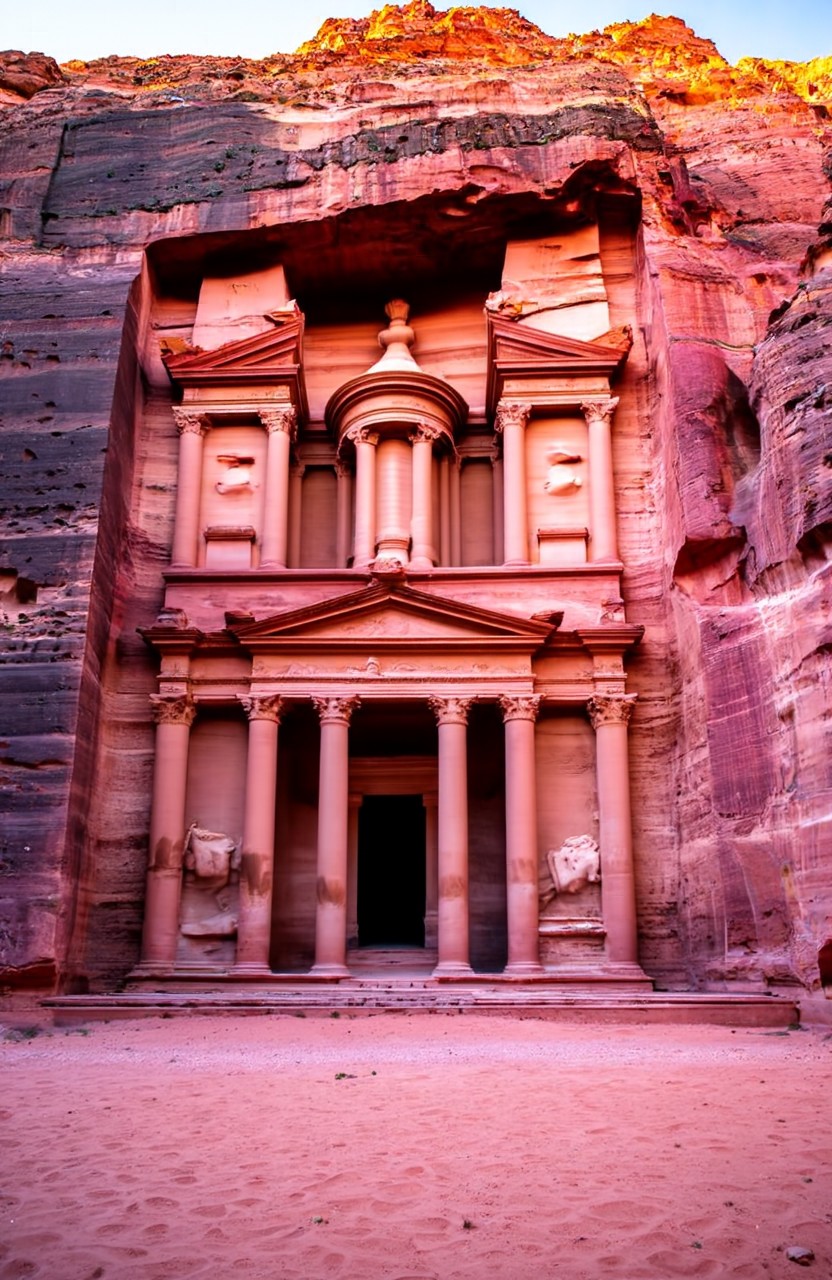}\\[1mm]{\raggedright\scriptsize\color{g600} The ancient city of Petra with the Treasury carved into pink sandstone, morning light\par}\end{minipage}\hspace{0.02\linewidth}\begin{minipage}[t]{0.3200\linewidth}\includegraphics[width=\linewidth]{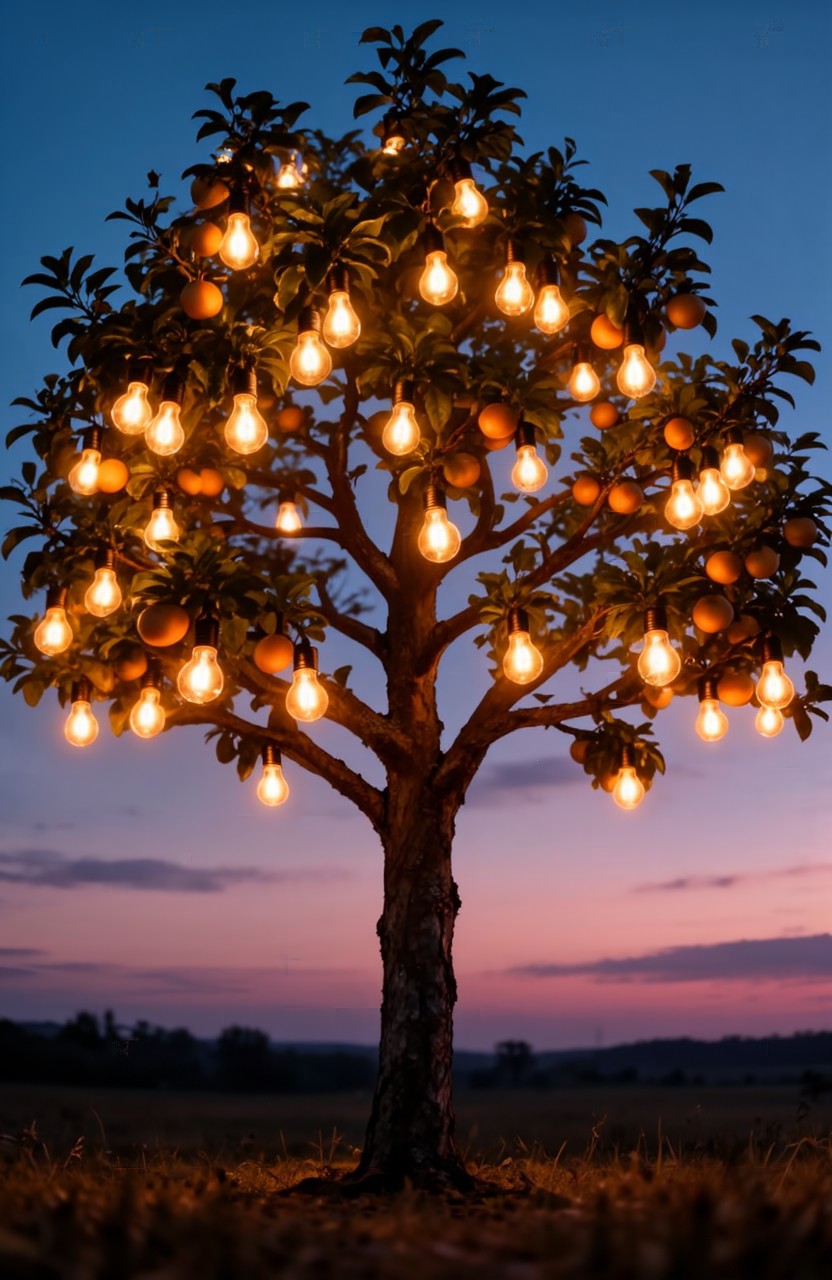}\\[1mm]{\raggedright\scriptsize\color{g600} A tree with lightbulbs instead of fruits glowing at dusk, surreal concept art\par}\end{minipage}\par
\vspace{4mm}
\noindent \begin{minipage}[t]{0.2350\linewidth}\includegraphics[width=\linewidth]{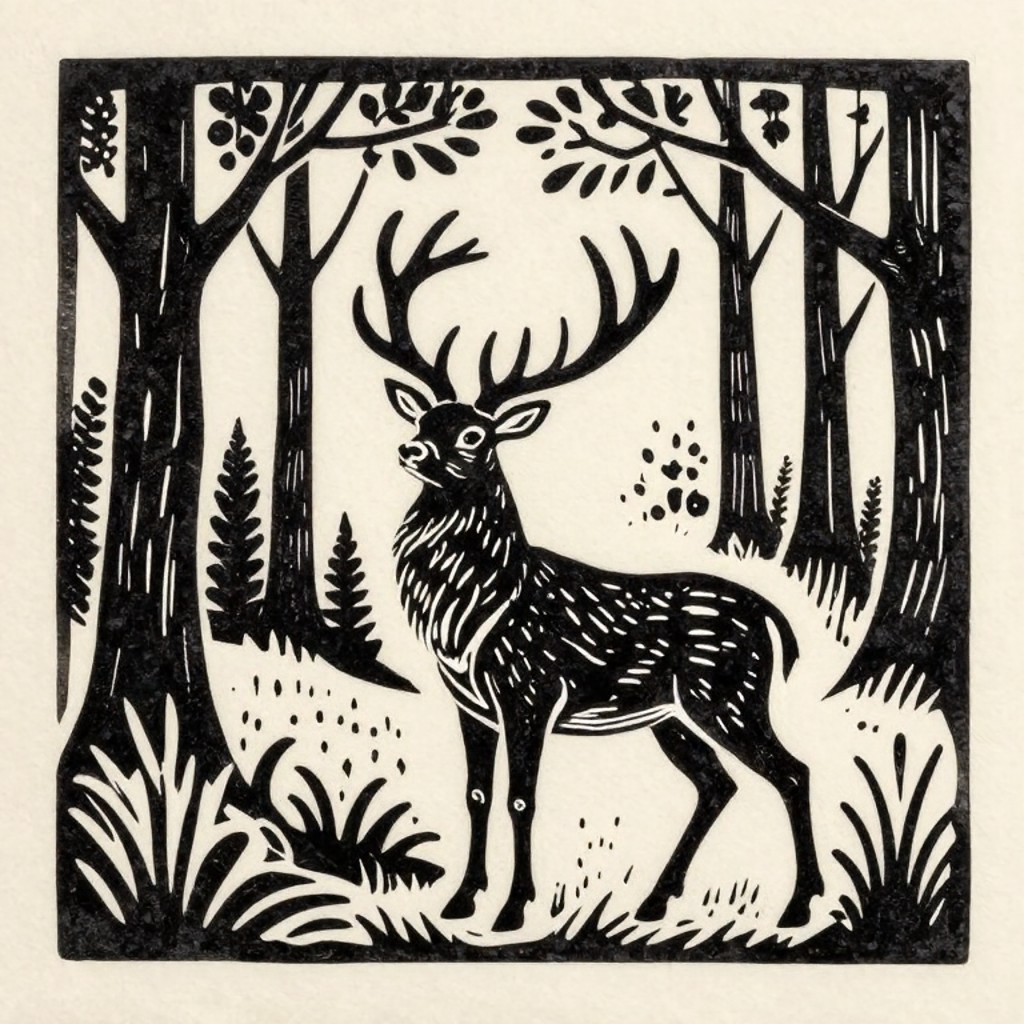}\\[1mm]{\raggedright\scriptsize\color{g600} A linocut print of a stag in a forest, bold black carved lines on cream paper, folk art style\par}\end{minipage}\hspace{0.02\linewidth}\begin{minipage}[t]{0.2350\linewidth}\includegraphics[width=\linewidth]{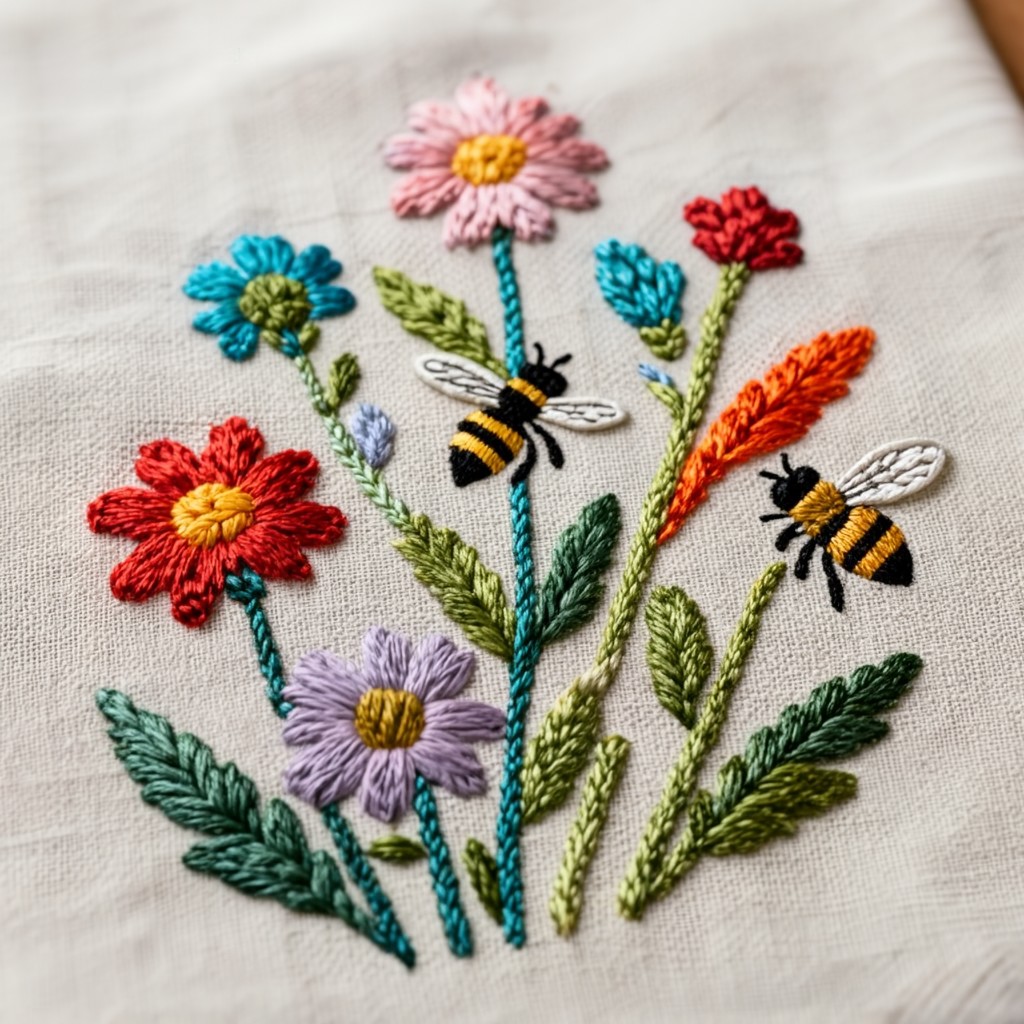}\\[1mm]{\raggedright\scriptsize\color{g600} An intricate embroidery of wildflowers and bees on linen fabric, colorful silk threads, close-up textile photograph\par}\end{minipage}\hspace{0.02\linewidth}\begin{minipage}[t]{0.2350\linewidth}\includegraphics[width=\linewidth]{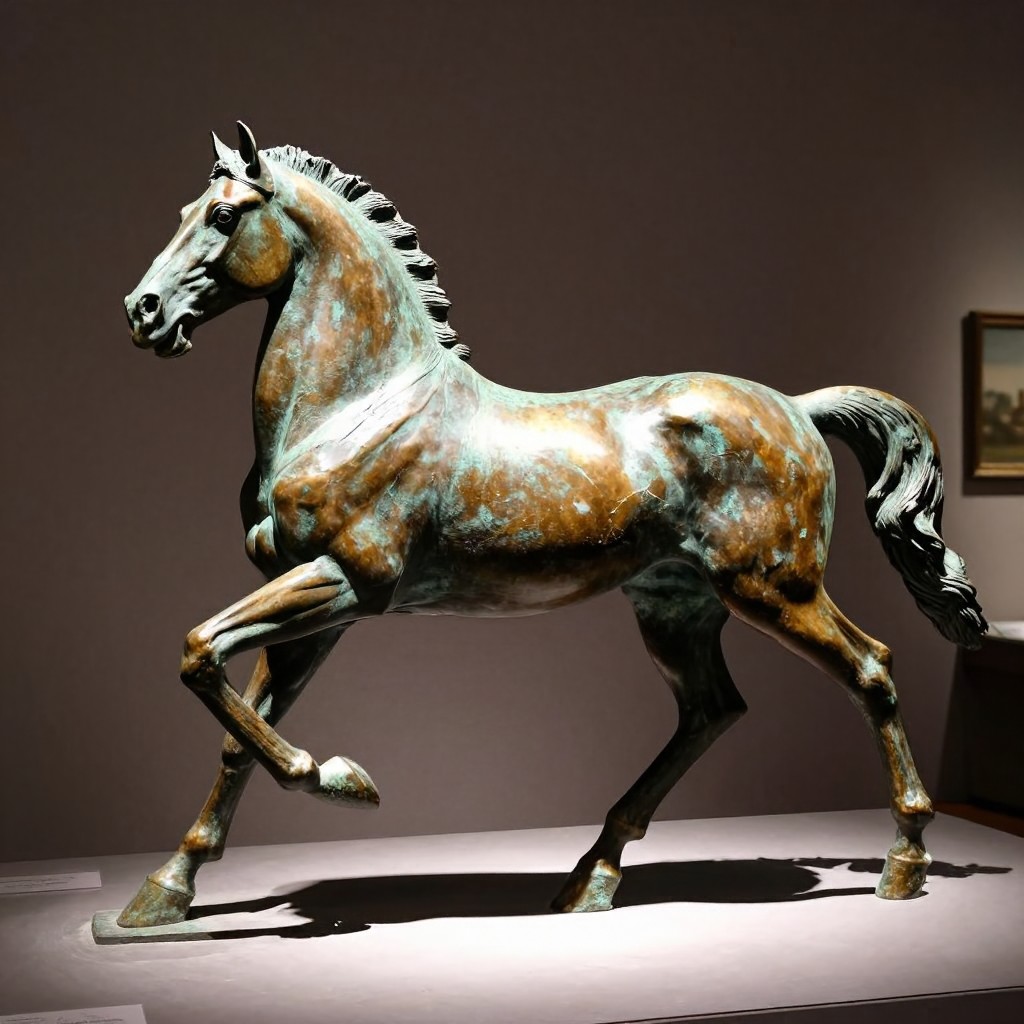}\\[1mm]{\raggedright\scriptsize\color{g600} A bronze sculpture of a horse in motion, patina, dramatic museum spotlight\par}\end{minipage}\hspace{0.02\linewidth}\begin{minipage}[t]{0.2350\linewidth}\includegraphics[width=\linewidth]{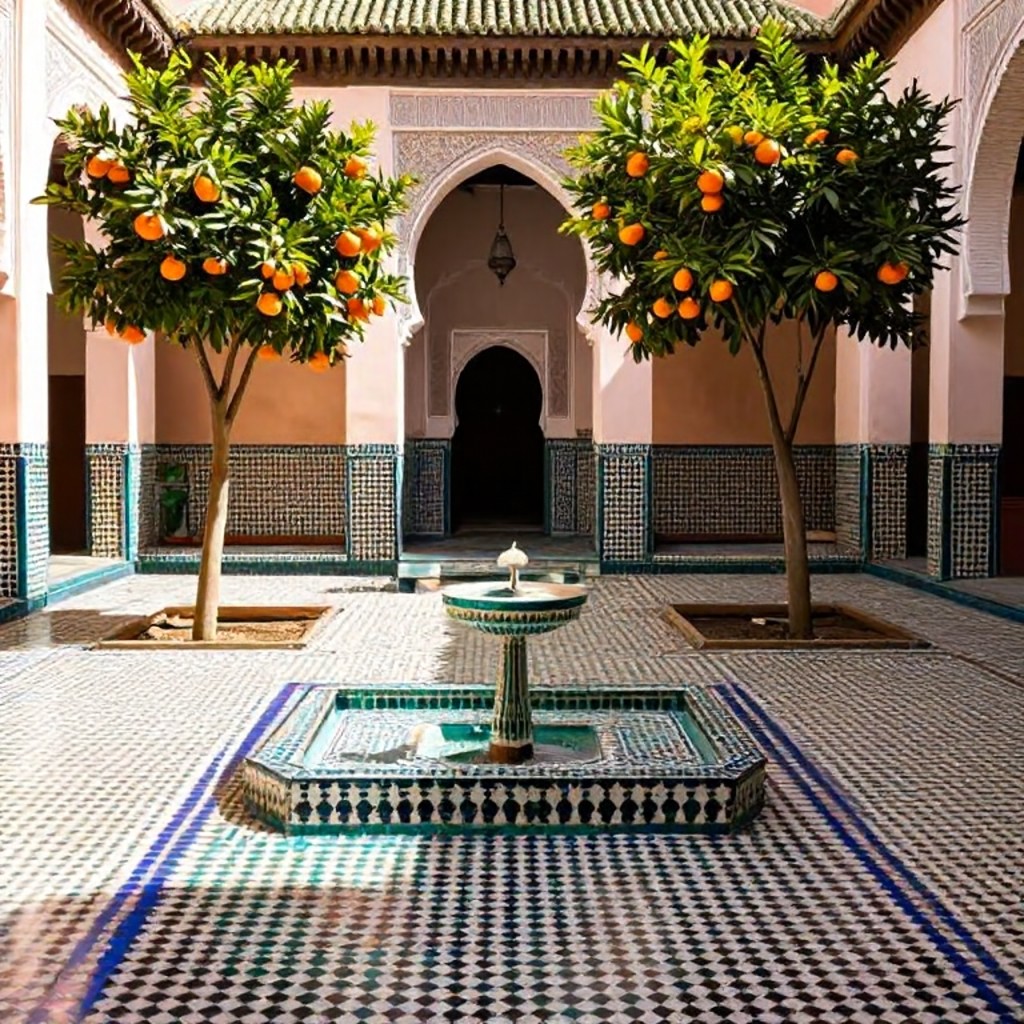}\\[1mm]{\raggedright\scriptsize\color{g600} A Moroccan riad courtyard with intricate zellige tiles, a fountain, orange trees and arched doorways, warm sunlight and long shadows\par}\end{minipage}\par

\clearpage
\section{Additional Depth Results}
\label{app:depth}
Predictions of the three arms of Tab.~\ref{tab:depth} on two evenly spaced images from each of
NYUv2, KITTI, ETH3D, ScanNet and DIODE (top to bottom), not selected by quality. Each panel shows the
raw relative-log-depth prediction as inverse depth (near is bright), scaled to its own 2nd--98th
percentile; no ground-truth alignment is applied.

\begin{figure}[H]\centering
\includegraphics[width=\linewidth,height=0.86\textheight,keepaspectratio]{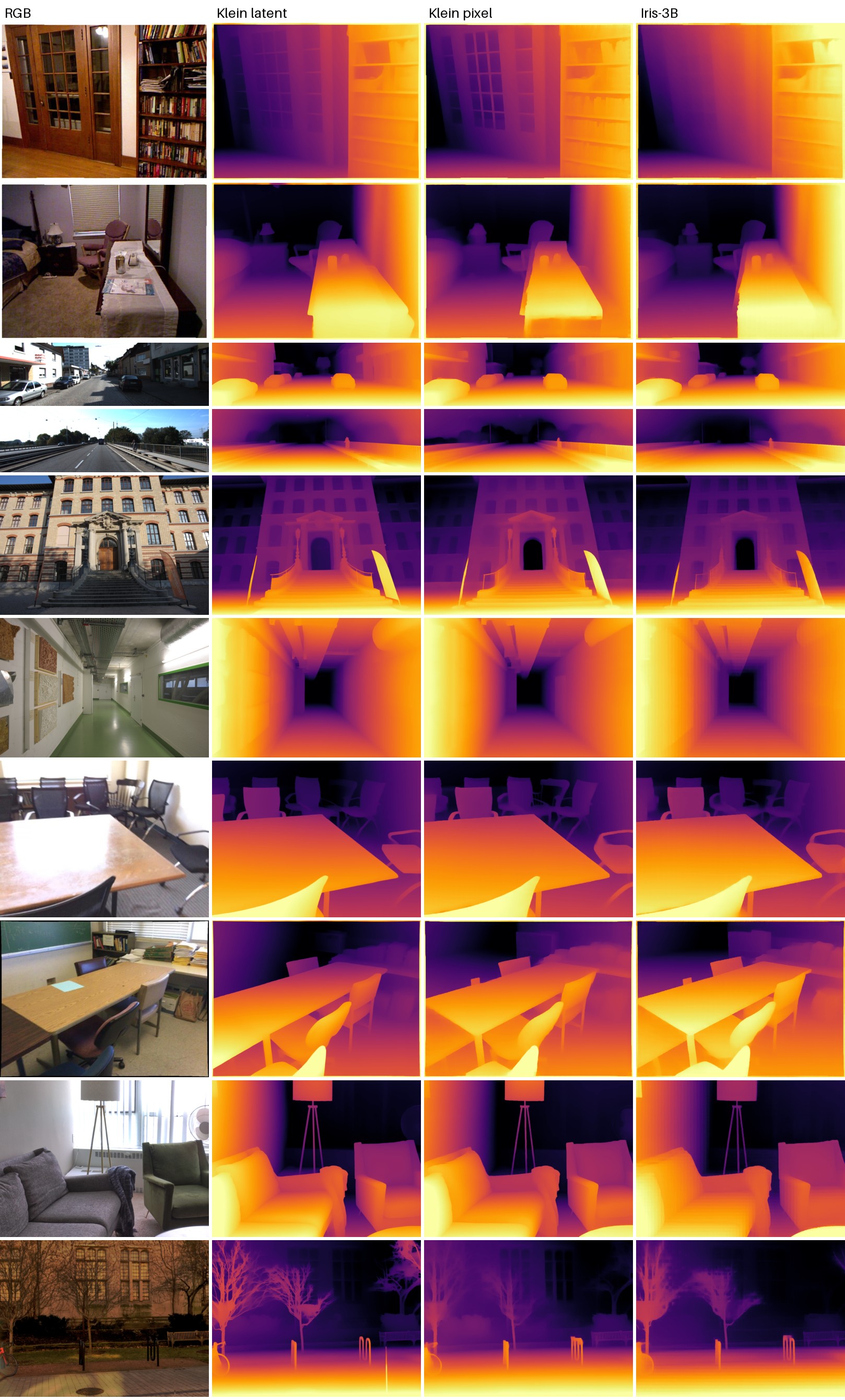}
\caption{Depth on the five zero-shot benchmarks after 10K steps of the same direct-regression recipe.}
\label{fig:depth-benchmarks}
\end{figure}

\clearpage
\section{Additional Restoration Results}
\label{app:sr-real}
All panels use the same models and settings as Fig.~\ref{fig:sr-div2k}: one forward pass and an
empty prompt, except that the Iris-3B panels of Fig.~\ref{fig:sr-reallq} come from an earlier
checkpoint of the same run (5K steps). We chose the images and regions after viewing the outputs.

\begin{figure}[H]\centering
\includegraphics[width=\linewidth]{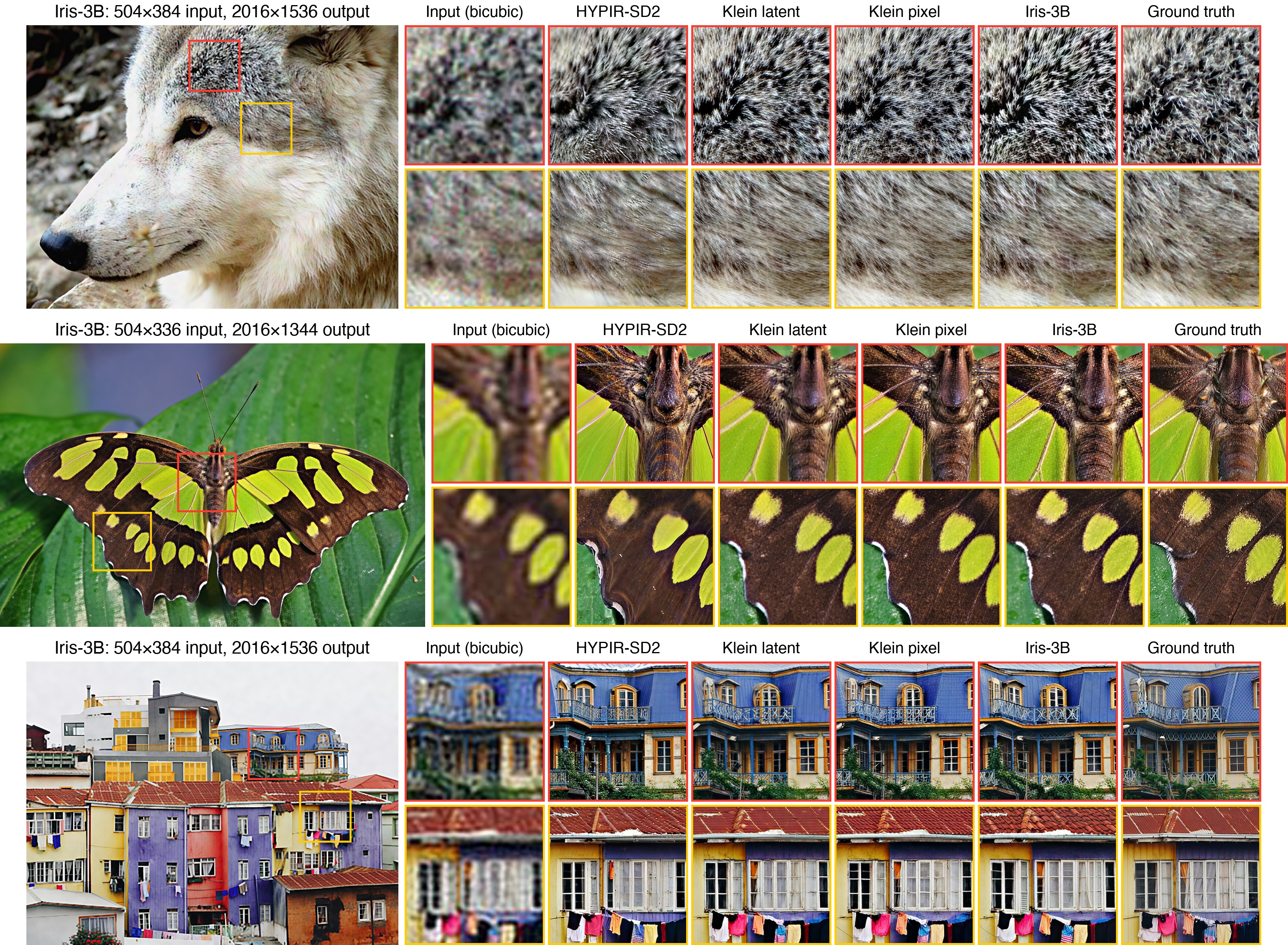}
\caption{$4\times$ restoration of further DIV2K validation images with native Real-ESRGAN
degradations (the Tab.~\ref{tab:sr} pairs).}
\label{fig:sr-div2k-extra1}
\end{figure}

\begin{figure}[H]\centering
\includegraphics[width=\linewidth]{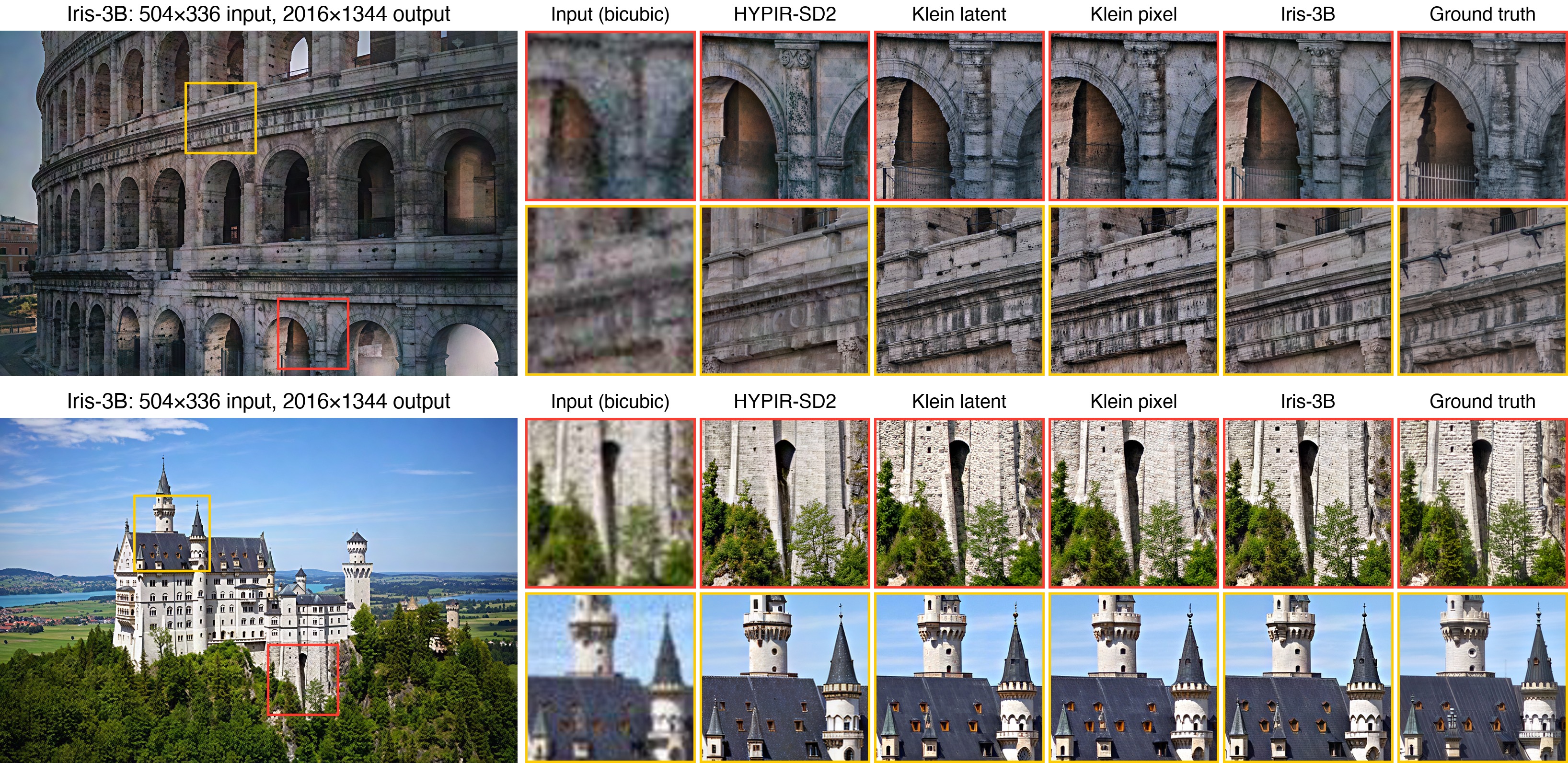}
\caption{$4\times$ restoration of further DIV2K validation images (continued).}
\label{fig:sr-div2k-extra2}
\end{figure}

\clearpage
\begin{figure}[H]\centering
\includegraphics[width=\linewidth,height=0.9\textheight,keepaspectratio]{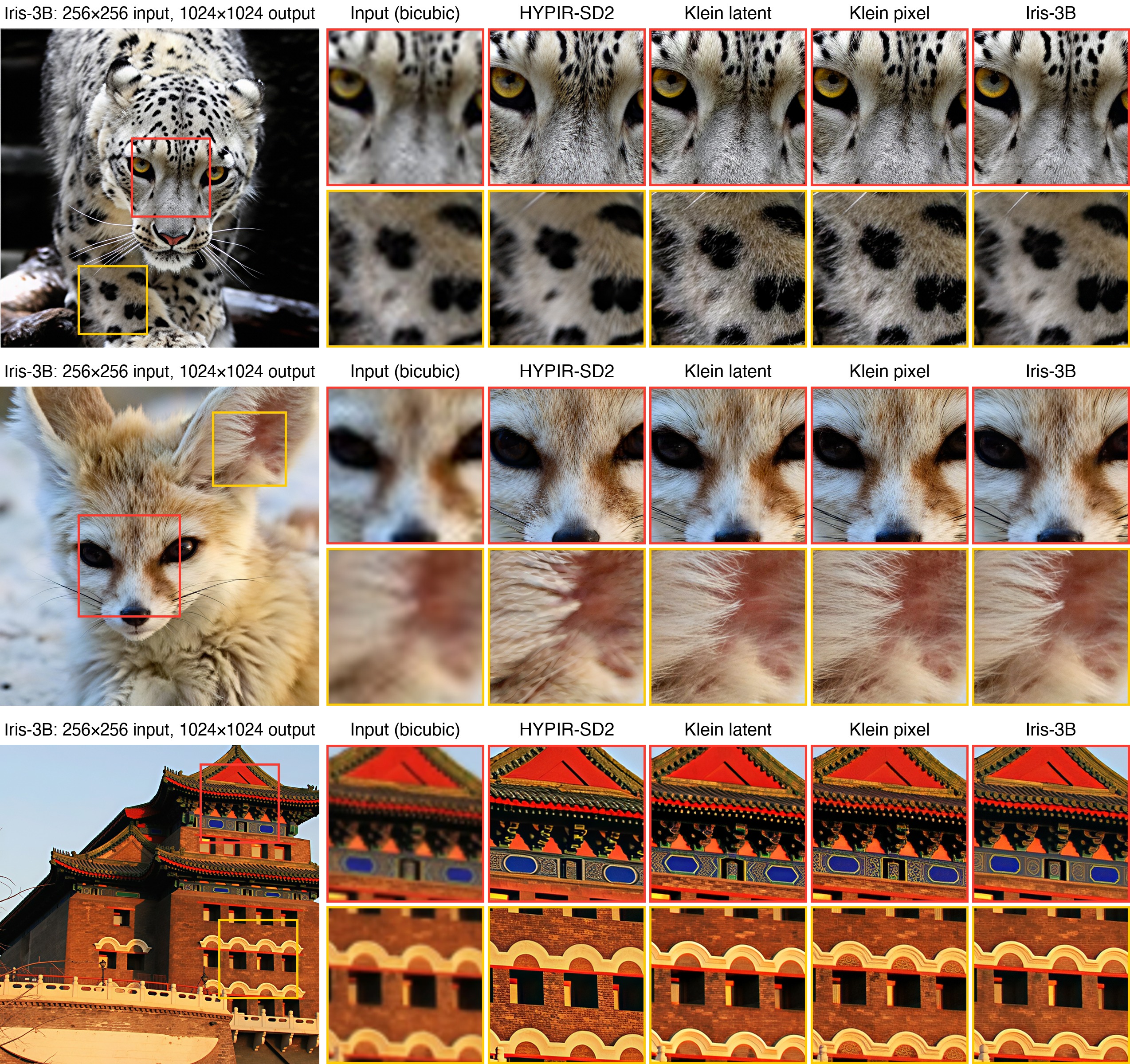}
\caption{$4\times$ restoration of real-world low-quality images from RealLQ250~\cite{dreamclear} (no ground truth).}
\label{fig:sr-reallq}
\end{figure}

\clearpage
\begin{figure}[H]\centering
\includegraphics[width=\linewidth,height=0.9\textheight,keepaspectratio]{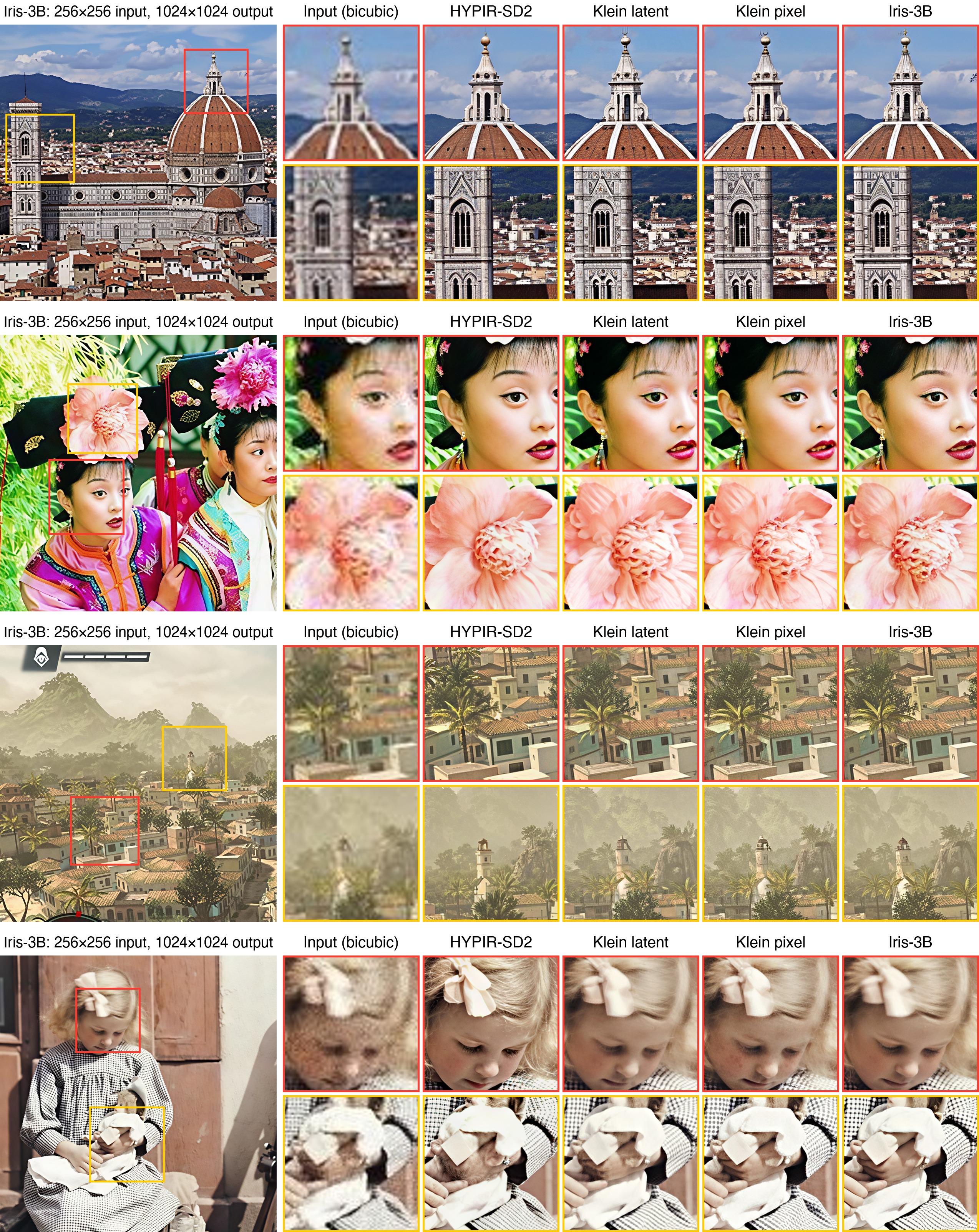}
\caption{$4\times$ restoration of four real-world example inputs from the public HYPIR
repository~\cite{hypir} (no ground truth).}
\label{fig:sr-real}
\end{figure}


\begin{thebibliography}{99}
\bibitem{flux2} Black Forest Labs.
\newblock FLUX.2 [klein]: Towards Interactive Visual Intelligence.
\newblock \url{https://bfl.ai/blog/flux2-klein-towards-interactive-visual-intelligence}, 2026.

\bibitem{pixeldit}
Yongsheng Yu, Wei Xiong, Weili Nie, Yichen Sheng, Shiqiu Liu, and Jiebo Luo.
\newblock PixelDiT: Pixel Diffusion Transformers for Image Generation.
\newblock arXiv:2511.20645, 2025.

\bibitem{qwenimage} Chenfei Wu et al.
\newblock Qwen-Image Technical Report.
\newblock arXiv:2508.02324, 2025.

\bibitem{ldm}
Robin Rombach, Andreas Blattmann, Dominik Lorenz, Patrick Esser, and Bj\"orn Ommer.
\newblock High-Resolution Image Synthesis with Latent Diffusion Models.
\newblock In CVPR, 2022.

\bibitem{jit}
Tianhong Li and Kaiming He.
\newblock Back to Basics: Let Denoising Generative Models Denoise.
\newblock arXiv:2511.13720, 2025.

\bibitem{asymflow}
Hansheng Chen, Jan Ackermann, Minseo Kim, Gordon Wetzstein, and Leonidas Guibas.
\newblock Asymmetric Flow Models.
\newblock arXiv:2605.12964, 2026.

\bibitem{l2p}
Zhennan Chen, Junwei Zhu, Xu Chen, Jiangning Zhang, Jiawei Chen, Zhuoqi Zeng, Wei Zhang, Chengjie Wang, Jian Yang, and Ying Tai.
\newblock L2P: Unlocking Latent Potential for Pixel Generation.
\newblock arXiv:2605.12013, 2026.

\bibitem{empiricalstudy}
Dengyang Jiang, Ruoyi Du, Zhennan Chen, Dongyang Liu, Zanyi Wang, Mingzhe Zheng, Xiangpeng Yang, Huanqia Cai, Aiming Hao, Yuming Jiang, Peng Gao, Harry Yang, and Steven Hoi.
\newblock An Empirical Study of Training Pixel-Space Text-to-Image Diffusion Models.
\newblock arXiv:2608.16887, 2026.

\bibitem{pixelumm}
Cong Wei, Xuanchi Ren, Bryan Chu, Weiming Ren, Huan Ling, Jiahui Huang, Laura Leal-Taix\'e, Sanja Fidler, Wenhu Chen, Zian Wang, and Jay Zhangjie Wu.
\newblock PixelUMM: Encoder-Free Unified Image and Video Understanding and Generation.
\newblock arXiv:2609.38597, 2026.

\bibitem{marigold}
Bingxin Ke, Anton Obukhov, Shengyu Huang, Nando Metzger, Rodrigo Caye Daudt, and Konrad Schindler.
\newblock Repurposing Diffusion-Based Image Generators for Monocular Depth Estimation.
\newblock In CVPR, 2024.

\bibitem{marigoldv2}
Igor Pavlovic, Thiemo Wandel, Anton Obukhov, Luca Bartolomei, Andrey Davydov, Fabio Tosi, Matteo Poggi, Sabine S\"usstrunk, and Dengxin Dai.
\newblock Marigold V2: Revisiting Diffusion Transformers for Monocular Depth Estimation.
\newblock arXiv:2609.08084, 2026.

\bibitem{hypir}
Xinqi Lin et al.
\newblock Harnessing Diffusion-Yielded Score Priors for Image Restoration.
\newblock arXiv:2507.20590, 2025.

\bibitem{dip} Zhennan Chen et al.
\newblock DiP: Taming Diffusion Models in Pixel Space.
\newblock arXiv:2511.18822, 2025.

\bibitem{pixnerd} Shuai Wang et al.
\newblock PixNerd: Pixel Neural Field Diffusion.
\newblock arXiv:2507.23268, 2025.

\bibitem{pixelgen} Zehong Ma, Ruihan Xu, and Shiliang Zhang.
\newblock PixelGen: Improving Pixel Diffusion with Perceptual Supervision.
\newblock arXiv:2602.02493, 2026.

\bibitem{e2eft} Gonzalo Martin Garcia et al.
\newblock Fine-Tuning Image-Conditional Diffusion Models is Easier than You Think.
\newblock In WACV, 2025.

\bibitem{pointdit}
Haofei Xu, Rundi Wu, Philipp Henzler, Nikolai Kalischek, Michael Oechsle, Fabian Manhardt, Marc Pollefeys, Andreas Geiger, Federico Tombari, and Michael Niemeyer.
\newblock PointDiT: Pixel-Space Diffusion for Monocular Geometry Estimation.
\newblock arXiv:2607.02515, 2026.

\bibitem{dinov3} Oriane Sim\'eoni et al.
\newblock DINOv3.
\newblock arXiv:2508.10104, 2025.

\bibitem{visionbanana}
Valentin Gabeur et al.
\newblock Image Generators are Generalist Vision Learners.
\newblock arXiv:2604.20329, 2026.

\bibitem{stablesr} Jianyi Wang, Zongsheng Yue, Shangchen Zhou, Kelvin C.~K. Chan, and Chen Change Loy.
\newblock Exploiting Diffusion Prior for Real-World Image Super-Resolution.
\newblock IJCV, 2024.

\bibitem{diffbir} Xinqi Lin et al.
\newblock DiffBIR: Towards Blind Image Restoration with Generative Diffusion Prior.
\newblock In ECCV, 2024.

\bibitem{seesr} Rongyuan Wu, Tao Yang, Lingchen Sun, Zhengqiang Zhang, Shuai Li, and Lei Zhang.
\newblock SeeSR: Towards Semantics-Aware Real-World Image Super-Resolution.
\newblock In CVPR, 2024.

\bibitem{supir} Fanghua Yu, Jinjin Gu, Zheyuan Li, Jinfan Hu, Xiangtao Kong, Xintao Wang, Jingwen He,
Yu Qiao, and Chao Dong.
\newblock Scaling Up to Excellence: Practicing Model Scaling for Photo-Realistic Image Restoration
In the Wild.
\newblock In CVPR, 2024.

\bibitem{osediff} Rongyuan Wu, Lingchen Sun, Zhiyuan Ma, and Lei Zhang.
\newblock One-Step Effective Diffusion Network for Real-World Image Super-Resolution.
\newblock In NeurIPS, 2024.

\bibitem{realesrgan} Xintao Wang, Liangbin Xie, Chao Dong, and Ying Shan.
\newblock Real-ESRGAN: Training Real-World Blind Super-Resolution with Pure Synthetic Data.
\newblock In ICCV Workshops, 2021.

\bibitem{repa} Sihyun Yu et al.
\newblock Representation Alignment for Generation: Training Diffusion Transformers Is Easier Than You Think.
\newblock In ICLR, 2025.

\bibitem{dinov2} Maxime Oquab et al.
\newblock DINOv2: Learning Robust Visual Features without Supervision.
\newblock TMLR, 2024.

\bibitem{irepa} Jaskirat Singh et al.
\newblock What Matters for Representation Alignment: Global Information or Spatial Structure?
\newblock arXiv:2512.10794, 2025.

\bibitem{selfflow} Hila Chefer, Patrick Esser, Dominik Lorenz, Dustin Podell, Vikash Raja, Vinh Tong, Antonio Torralba, and Robin Rombach.
\newblock Self-Supervised Flow Matching for Scalable Multi-Modal Synthesis.
\newblock In ICML, 2026.

\bibitem{geneval} Dhruba Ghosh, Hannaneh Hajishirzi, and Ludwig Schmidt.
\newblock GenEval: An Object-Focused Framework for Evaluating Text-to-Image Alignment.
\newblock In NeurIPS Datasets and Benchmarks, 2023.

\bibitem{ella} Xiwei Hu et al.
\newblock ELLA: Equip Diffusion Models with LLM for Enhanced Semantic Alignment.
\newblock arXiv:2403.05135, 2024.

\bibitem{flowmatching} Yaron Lipman, Ricky T.~Q. Chen, Heli Ben-Hamu, Maximilian Nickel, and Matt Le.
\newblock Flow Matching for Generative Modeling.
\newblock In ICLR, 2023.

\bibitem{flux1} Black Forest Labs.
\newblock FLUX.1.
\newblock \url{https://github.com/black-forest-labs/flux}, 2024.

\bibitem{sd3} Patrick Esser et al.
\newblock Scaling Rectified Flow Transformers for High-Resolution Image Synthesis.
\newblock In ICML, 2024.

\bibitem{gqa} Joshua Ainslie, James Lee-Thorp, Michiel de Jong, Yury Zemlyanskiy, Federico Lebr\'on,
and Sumit Sanghai.
\newblock GQA: Training Generalized Multi-Query Transformer Models from Multi-Head Checkpoints.
\newblock In EMNLP, 2023.

\bibitem{gatedattn} Zihan Qiu et al.
\newblock Gated Attention for Large Language Models: Non-linearity, Sparsity, and Attention-Sink-Free.
\newblock In NeurIPS, 2025.

\bibitem{cogview} Ming Ding et al.
\newblock CogView: Mastering Text-to-Image Generation via Transformers.
\newblock In NeurIPS, 2021.

\bibitem{rmsnorm} Biao Zhang and Rico Sennrich.
\newblock Root Mean Square Layer Normalization.
\newblock In NeurIPS, 2019.

\bibitem{zimage} Z-Image Team.
\newblock Z-Image: An Efficient Image Generation Foundation Model with Single-Stream Diffusion Transformer.
\newblock arXiv:2511.22699, 2025.

\bibitem{krea2} Krea.
\newblock Krea 2 Technical Report.
\newblock \url{https://www.krea.ai/blog/krea-2-technical-report}, 2026.

\bibitem{dit} William Peebles and Saining Xie.
\newblock Scalable Diffusion Models with Transformers.
\newblock In ICCV, 2023.

\bibitem{pixartalpha} Junsong Chen et al.
\newblock PixArt-$\alpha$: Fast Training of Diffusion Transformer for Photorealistic Text-to-Image
Synthesis.
\newblock In ICLR, 2024.

\bibitem{qwen3vl} Shuai Bai et al.
\newblock Qwen3-VL Technical Report.
\newblock arXiv:2511.21631, 2025.

\bibitem{unifusion} Kevin Li, Manuel Brack, Sudeep Katakol, Hareesh Ravi, and Ajinkya Kale.
\newblock UniFusion: Vision-Language Model as Unified Encoder in Image Generation.
\newblock arXiv:2510.12789, 2025.

\bibitem{rope} Jianlin Su, Yu Lu, Shengfeng Pan, Ahmed Murtadha, Bo Wen, and Yunfeng Liu.
\newblock RoFormer: Enhanced Transformer with Rotary Position Embedding.
\newblock Neurocomputing, 2024.

\bibitem{ropevit} Byeongho Heo, Song Park, Dongyoon Han, and Sangdoo Yun.
\newblock Rotary Position Embedding for Vision Transformer.
\newblock In ECCV, 2024.

\bibitem{muonblog} Keller Jordan et al.
\newblock Muon: An Optimizer for Hidden Layers in Neural Networks.
\newblock \url{https://kellerjordan.github.io/posts/muon/}, 2024.

\bibitem{muon} Jingyuan Liu et al.
\newblock Muon is Scalable for LLM Training.
\newblock arXiv:2502.16982, 2025.

\bibitem{xomni} Zigang Geng et al.
\newblock X-Omni: Reinforcement Learning Makes Discrete Autoregressive Image Generative Models Great Again.
\newblock arXiv:2507.22058, 2025.

\bibitem{oneig} Jingjing Chang et al.
\newblock OneIG-Bench: Omni-dimensional Nuanced Evaluation for Image Generation.
\newblock arXiv:2506.07977, 2025.

\bibitem{i1} Boya Zeng et al.
\newblock i1: A Simple and Fully Open Recipe for Strong Text-to-Image Models.
\newblock arXiv:2606.11289, 2026.

\bibitem{div2k} Eirikur Agustsson and Radu Timofte.
\newblock NTIRE 2017 Challenge on Single Image Super-Resolution: Dataset and Study.
\newblock In CVPR Workshops, 2017.

\bibitem{invsr} Zongsheng Yue, Kang Liao, and Chen Change Loy.
\newblock Arbitrary-steps Image Super-resolution via Diffusion Inversion.
\newblock In CVPR, 2025.

\bibitem{s3diff} Aiping Zhang, Zongsheng Yue, Renjing Pei, Wenqi Ren, and Xiaochun Cao.
\newblock Degradation-Guided One-Step Image Super-Resolution with Diffusion Priors.
\newblock arXiv:2409.17058, 2024.

\bibitem{dreamclear} Yuang Ai et al.
\newblock DreamClear: High-Capacity Real-World Image Restoration with Privacy-Safe Dataset Curation.
\newblock In NeurIPS, 2024.
\end{thebibliography}
\end{document}